\documentclass{article} % For LaTeX2e
\usepackage[preprint]{colm2026_conference}

\usepackage{microtype}
\usepackage{hyperref}
\usepackage{url}
\usepackage{booktabs}
\usepackage{multirow}
\usepackage{graphicx}
\usepackage{amsmath,amssymb}
\usepackage[table]{xcolor}
\usepackage{caption}
\usepackage{pifont}
\newcommand{\cmark}{\ding{51}}
\newcommand{\xmark}{\ding{55}}
\usepackage{algorithm}
\usepackage{algpseudocode}
\usepackage{subcaption}  % preamble
\usepackage{wrapfig}
\usepackage{tcolorbox}
\tcbuselibrary{skins, breakable}

\usepackage{titlesec}
\titlespacing*{\section}{0pt}{0.8ex plus 0.2ex minus 0.2ex}{0.4ex plus 0.1ex}
\titlespacing*{\subsection}{0pt}{0.6ex plus 0.2ex minus 0.2ex}{0.3ex plus 0.1ex}

\definecolor{promptframe1}{HTML}{927AB8}
\definecolor{promptframe2}{HTML}{5A87B8}
\newtcolorbox{promptbox}[1]{
  enhanced,
  colback=promptframe1!5,
  colframe=promptframe1,
  coltitle=white,
  fonttitle=\bfseries,
  title=#1,
  arc=1.5mm,
  boxrule=0.8pt,
  left=6pt, right=6pt, top=4pt, bottom=4pt,
}
\newtcolorbox{promptbox2}[1]{
  enhanced,
  colback=promptframe2!5,
  colframe=promptframe2,
  coltitle=white,
  fonttitle=\bfseries,
  title=#1,
  arc=1.5mm,
  boxrule=0.8pt,
  left=6pt, right=6pt, top=4pt, bottom=4pt,
}
\usepackage{lineno}

\definecolor{darkblue}{rgb}{0, 0, 0.5}
\hypersetup{colorlinks=true, citecolor=darkblue, linkcolor=darkblue, urlcolor=darkblue}

\title{\textsc{EvolveRouter}: Co-Evolving Routing and Prompt for Multi-Agent Question Answering}

\author{
    Jiatan Huang\textsuperscript{1*}, Zheyuan Zhang\textsuperscript{2*}, Kaiwen Shi\textsuperscript{2}, Yanfang Ye\textsuperscript{2}, Chuxu Zhang\textsuperscript{1,$\dag$} \\
    \textsuperscript{1}University of Connecticut, USA~\textsuperscript{2}University of Notre Dame, USA\\
    \{jiatan.huang, chuxu.zhang\}@uconn.edu, \{zzhang42, yye7\}@nd.edu\\
\textsuperscript{*}Equal Contribution
\textsuperscript{$\dag$}Corresponding author
}
\begin{document}

\ifcolmsubmission
\linenumbers
\fi

\maketitle

\begin{abstract}
Large language model agents often exhibit complementary strengths, making routing a promising approach for multi-agent question answering. However, existing routing methods remain limited in two important ways: they typically optimize over a fixed pool of agents without improving the agents themselves, and they often rely on rigid collaboration schemes that cannot adapt the number of participating agents to the query. We propose \textsc{EvolveRouter}, a trainable framework that addresses both limitations by jointly improving agent quality and collaboration structure. First, EvolveRouter couples graph-based query routing with targeted instruction refinement in a closed-loop co-evolution process, allowing router diagnostics to guide agent improvement while refined agents provide cleaner supervision for routing. Second, it introduces an adaptive inference strategy that dynamically determines the effective collaboration size for each query through router-weighted answer agreement. Together, these designs enable more capable and more efficient multi-agent reasoning. Experiments on five question answering benchmarks show that EvolveRouter consistently outperforms SOTA routing baselines in both F1 and exact match, while further analysis confirms the benefits of closed-loop refinement and adaptive collaboration. Our code repo is available \href{https://anonymous.4open.science/r/EvolveRouter-3E94}{here}.
\end{abstract}

% \begin{abstract}
% Multi-agent question answering benefits from routing, which selects and weights agents for each query. Existing routing methods, however, operate over a fixed agent pool. As a result, they improve how agents are coordinated without improving the agents themselves. This limits routing effectiveness, since the router is ultimately constrained by the quality of the agents it routes over. We propose \textsc{EvolveRouter}, a framework built on the observation that routing quality and agent quality can improve each other: a well-trained router reveals which agents fail and why, and improving those agents in turn gives the router better candidates to work with. Concretely, the framework couples graph-based routing with iterative prompt refinement. The router is first trained while collecting diagnostic evidence about agent failures. These diagnostics are then used to identify underperforming but important agents, revise their prompts, and retrain the router on the updated agent pool. In this way, routing and agent quality improve together in a closed loop. We further introduce an answer-agreement stopping rule that adaptively determines how many agents to query at test time. Results on five question answering benchmarks show that EvolveRouter consistently outperforms strong routing baselines in both F1 and exact match while improving inference efficiency.
% \end{abstract}

\section{Introduction}

\begin{figure*}[t]
\centering

\begin{subfigure}[t]{0.48\textwidth}
    \centering
    \includegraphics[width=\textwidth]{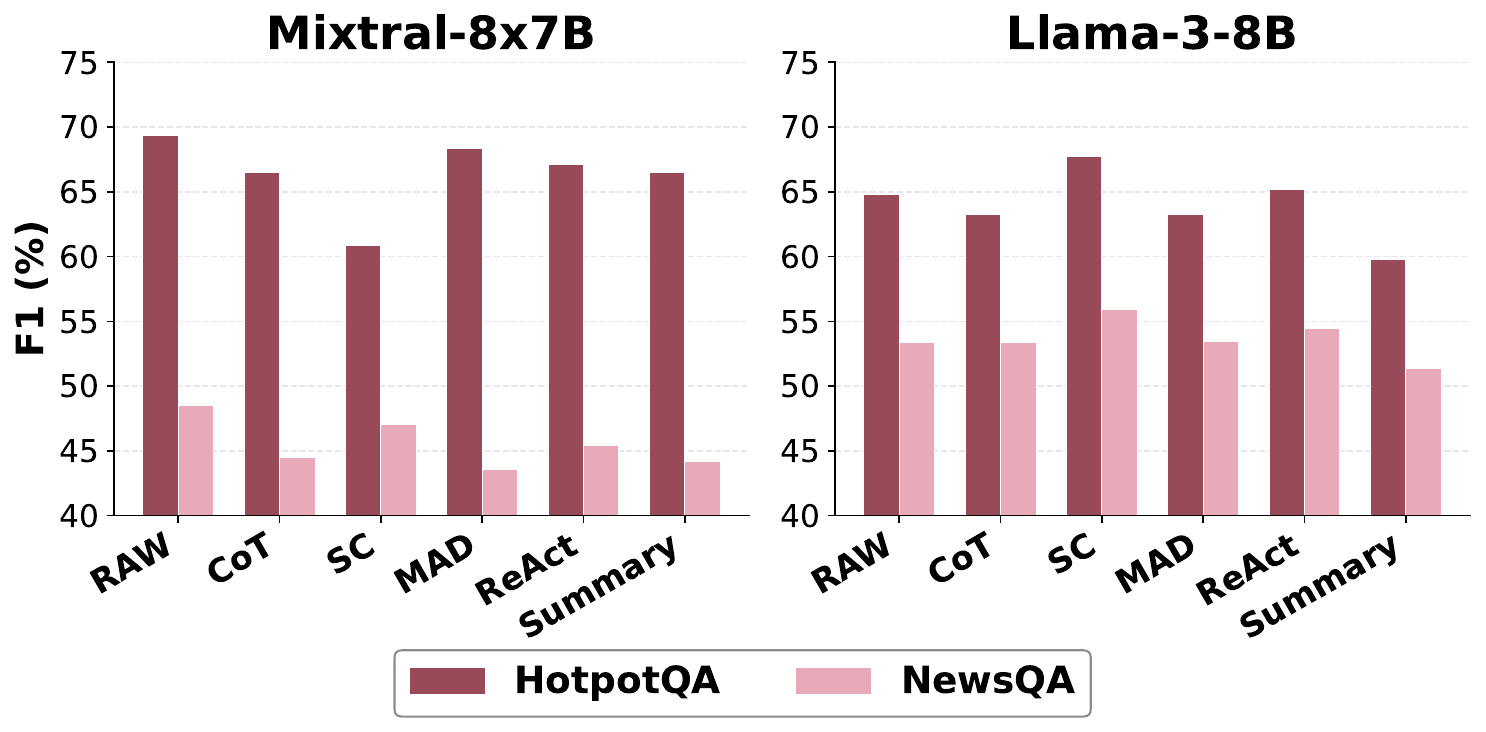}
    \vspace{-15pt}
    \caption{Same LLM, same prompt, different tasks. Performance varies across agent roles, with no single best options across tasks.}
    \label{fig:agent-task}
\end{subfigure}
\hfill
\begin{subfigure}[t]{0.48\textwidth}
    \centering
    \includegraphics[width=\textwidth]{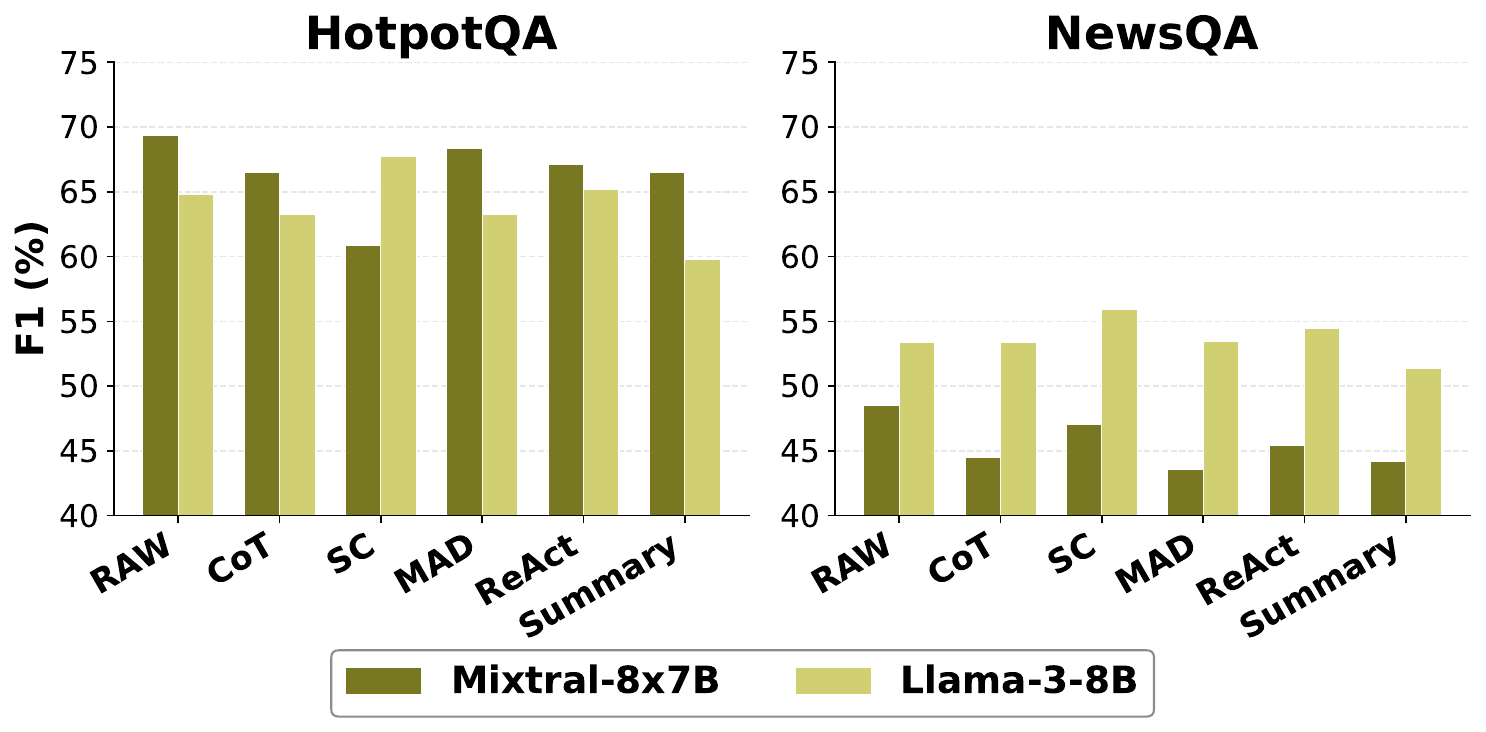}
    \vspace{-15pt}
    \caption{Same task, same prompt, different LLMs. The LLM-role performance gap varies by role, suggesting non-trivial interaction.}
    \label{fig:agent-backbone}
\end{subfigure}

\vspace{0.5em}

\begin{subfigure}[t]{\textwidth}
    \centering
    \includegraphics[width=\textwidth]{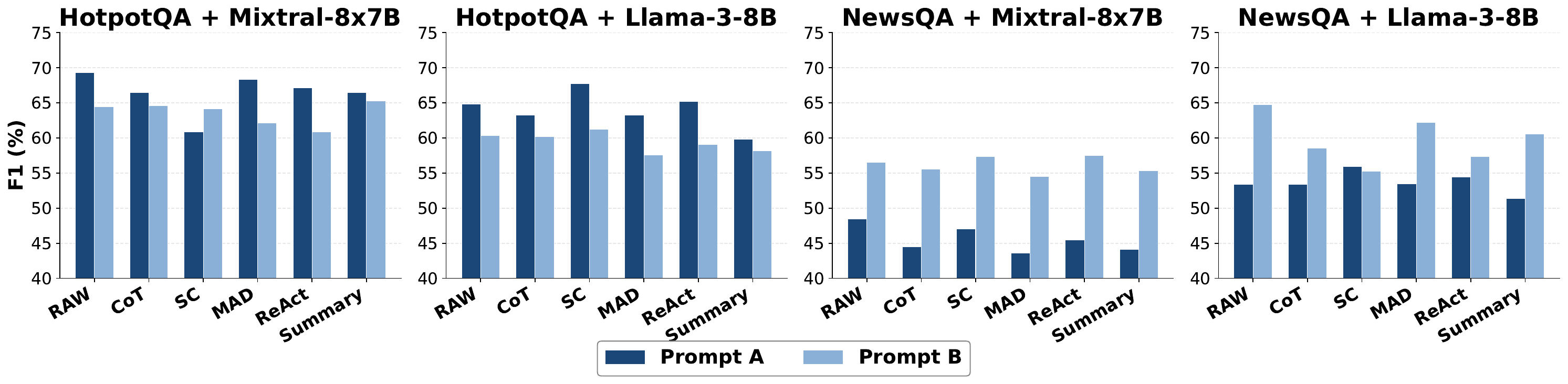}
    \vspace{-15pt}
    \caption{Same LLM and task, different prompts. Prompt choice substantially affects per-agent F1, with the effect varying across roles.}
    \label{fig:agent-prompt}
\end{subfigure}
\vspace{-5pt}
\caption{\textbf{Motivating observation: agent quality is highly context-dependent, and no single agent configuration is consistently optimal.} Multi-dimensional analysis of per-agent F1 (\%) across six agent roles, evaluated along three controlled dimensions: task variation (a), backbone variation (b), and prompt variation (c).}
\vspace{-25pt}
\label{fig:agent-comparison}
\end{figure*}

Recent years have witnessed remarkable progress in large language models (LLMs), which have demonstrated strong capabilities in reasoning, language understanding, and text generation~\citep{kojima2023largelanguagemodelszeroshot, li2025crochetbench, li2025adaptive}. Building on these advances, LLM-based agents have emerged as a powerful paradigm for autonomous problem solving, enabling models to plan, invoke tools, and execute multi-step tasks with limited human intervention~\citep{jimenez2024swebenchlanguagemodelsresolve, guo2024dsagentautomateddatascience, li2025embodiedagentinterfacebenchmarking}. As a result, LLM agents are increasingly being deployed in practical domains such as software engineering, data analysis, and interactive decision-making \citep{ma2025autodata}. However, this growing ecosystem also introduces a fundamental systems challenge: for a given downstream task, practitioners must choose from a large design space of model backbones, prompting strategies, and interaction protocols, while the most effective configuration is often unclear in advance. Prior work suggests that different agents often exhibit complementary, task-specific strengths, and within practical deployment regimes, moderately sized models or simpler agent configurations can sometimes match or even outperform more complex alternatives depending on the task setting (Also demonstrated empirically in Figure~\ref{fig:agent-comparison}) ~\citep{chen2023agentversefacilitatingmultiagentcollaboration, chen2024reconcileroundtableconferenceimproves}. At the same time, naively combining all available agents is inefficient: uniform ensembles often waste computation on irrelevant participants, while poorly selected multi-agent collaboration can even degrade final performance~\citep{wang2024mixtureofagentsenhanceslargelanguage}.

To address this challenge, a growing body of work has explored LLM routing and model selection as a promising direction. Early approaches relied on heuristic aggregation strategies such as majority voting and consistency-based selection ~\citep{wang2023selfconsistencyimproveschainthought, jiang-etal-2023-llm}, while later methods introduced adaptive routing that learns query-specific assignment policies through techniques such as preference learning and contrastive learning \citep{ong2025routellmlearningroutellms, wang2025mixllmdynamicroutingmixed, yue2025masrouterlearningroutellms}. More recent studies further leverage semantic and structural information from queries and context, moving beyond single-model selection toward collaborative multi-agent voting, where multiple agents are selected to jointly solve tasks in which individual agent alone achieves only suboptimal results ~\citep{feng2025graphroutergraphbasedrouterllm, zhang2025agentrouterknowledgegraphguidedllmrouter}.

Despite this progress, several important limitations remain. \textbf{First, most prior routing methods treat agent behavior as fixed, optimizing only \emph{which} agent to invoke rather than \emph{how well} each agent is configured to perform.} Prior work has shown that even minor instruction variations can induce substantial performance differences~\citep{sclar2024quantifyinglanguagemodelssensitivity, razavi2025benchmarkingpromptsensitivitylarge}, especially in moderate- and small-scale agent settings, as also illustrated in Figure~\ref{fig:agent-prompt}. As a result, routing alone cannot fully unlock collaboration gains when the underlying agent configurations remain suboptimal.

\textbf{Second, existing collaboration paradigms often rely on rigid participation schemes: they either route each query to a single agent or aggregate over a fixed set of agents.} The former risks missing complementary expertise, while the latter incurs unnecessary computation and may even hurt performance when irrelevant or weak agents are included. As a result, prior methods often fail to learn a truly adaptive collaboration structure that determines not only \emph{which} agents should participate, but also \emph{how many} and \emph{under what specialization} they should collaborate for each query.

To bridge these gaps, we propose \textsc{EvolveRouter}, a trainable framework that couples per-query routing with targeted agent instruction refinement through a closed-loop co-evolution process. \textbf{To address the first limitation}, our framework builds on prior routing work by first training a KG-based router using RouterGNN. During training, we collect fine-grained diagnostic signals on each agent’s strengths and failure patterns to identify consistently underperforming roles, generate candidate instruction revisions, and retain only those that lead to reliable improvements. The refined agents, in turn, produce higher-quality outputs and provide cleaner supervision for subsequent router training. Through this iterative process, routing and agent quality are improved jointly rather than in isolation. \textbf{To address the second limitation},  the rigid collaboration scheme is replaced with an adaptive inference strategy that dynamically determines how many agents should participate for each query. Instead of either selecting a single agent or aggregating over a fixed agent pool, our method invokes agents sequentially according to the router’s predicted ranking and monitors the weighted agreement among their answers. This enables dynamic estimation of the effective collaboration size \(K\) per query through router-weighted agreement, allowing our method to jointly learn both agent selection and collaboration scale.
Our contributions can be summarized as follows:
\begin{itemize}
    \vspace{-4pt}
    \item \textbf{Closed-loop co-evolution of routing and agent specialization.} To address the limitation of static agent behavior, we propose a trainable framework that couples query-aware routing with targeted agent instruction refinement. By iteratively using router diagnostics to improve underperforming agents, and refined agents to provide cleaner supervision for routing, EvolveRouter jointly improves both routing quality and agent capability.
    \vspace{-4pt}
    \item \textbf{Adaptive collaboration via dynamic agent selection.} To overcome rigid collaboration schemes, we introduce an adaptive inference mechanism that dynamically determines the effective collaboration size for each query. Rather than relying on either single-agent routing or fixed-pool aggregation, our method sequentially invokes router-ranked agents and estimates the required number of participants through router-weighted answer agreement.
    \vspace{-5pt}
    \item \textbf{Comprehensive experiments and analysis.} Extensive experiments on five QA benchmarks show that EvolveRouter consistently outperforms strong routing baselines. Further analysis validates the effectiveness of both closed-loop instruction refinement and adaptive collaboration, and provides insight into how routing quality and agent quality reinforce one another.
    \vspace{-5pt}
\end{itemize}

\section{Problem Formulation}
\label{sec:pf}

\subsection{Setup and Routing Objective}
\label{subsec:setup}
We consider an agent pool $\mathcal{A} = \{a_1, \ldots, a_n\}$, where each agent couples a backbone LLM with a prompting role such as chain-of-thought, debate, or react-reflect. In our setting, $n = 24$, corresponding to four backbones crossed with six roles. Each agent $a$ carries a prompt $\pi_a$. Given a query $q$ with context $\mathcal{C}$, the agent produces an answer $y_a(q)$ by calling its backbone LLM with its prompt. Since no single agent dominates across all queries (Figure~\ref{fig:agent-comparison}), the system benefits from adaptively selecting and weighting agents per query.

\begin{wraptable}{r}{0.5\textwidth}
\vspace{-12pt}
\centering
\small
\caption{Comparison of knowledge graph statistics between multi-hop, direct and domain-specific QA benchmarks. Agents and query node number are fixed across benchmarks.}
\vspace{-5pt}
\resizebox{\linewidth}{!}{
\begin{tabular}{lccc}
\toprule
\textbf{Statistics} & \textbf{HotpotQA} & \textbf{NewsQA} & \textbf{NGQA} \\
\midrule
\rowcolor{gray!20}
\multicolumn{4}{c}{\textit{Avg. \# of Nodes}} \\
Query   & 1    & 1    & 1    \\
Agent   & 24   & 24   & 24   \\
Entity  & 142.1 & 60.3 & 25.0 \\
\midrule
\rowcolor{gray!20}
\multicolumn{4}{c}{\textit{Avg. \# of Edges}} \\
Entity--Entity  & 385.1 & 109.4 & 26.75  \\
Agent--Entity   & 123.7 & 142.7 & 123.8 \\
Query--Entity   & 48.53  & 6.97   & 3.85   \\
\bottomrule
\end{tabular}
}
\label{tab:graph-stats}
\vspace{-10pt}
\end{wraptable}

\noindent\textbf{Routing objective.}
The goal is to learn a query-dependent distribution $p_\theta(a \mid q)$ over agents that reflects each agent's suitability for a given query. A key challenge is that suitability depends not only on the query and the agents, but also on the contextual entities and their relations, since these often determine which reasoning strategies are most effective. To capture this structure, we represent each QA instance as a typed knowledge graph $\mathcal{G} = (\mathcal{V}, \mathcal{E})$ with node set $\mathcal{V} = \mathcal{V}_Q \cup \mathcal{V}_A \cup \mathcal{V}_E$, where $\mathcal{V}_Q$, $\mathcal{V}_A$, and $\mathcal{V}_E$ contain query, agent, and entity nodes respectively. Edges encode four types of relationships: query--entity and entity--entity edges ground the query in its evidence context; agent--entity edges reflect each agent's perspective on the context; and query--agent edges are left as \emph{trainable} connections that carry the routing signal the model learns to predict. 
Within this formulation, routing reduces to learning a scoring function over the graph (construction details in Section~\ref{subsec:diag}):
    $s(q, a) = f_\theta(q, a;\, \mathcal{G})\,,\quad
    p_\theta(a \mid q, \mathcal{G})
      = \mathrm{softmax}_{a \in \mathcal{A}}\!\bigl(s(q, a)\bigr)$.

% \begin{table}[t]
% \centering
% \caption{Comparison of knowledge graph statistics across HotpotQA, NewsQA, and NGQA benchmarks. Agent and query node numbers are fixed across benchmarks.}
% \label{tab:graph-stats}
% \resizebox{0.45\columnwidth}{!}{
% \begin{tabular}{lccc}
% \toprule
% \textbf{Statistics} & \textbf{HotpotQA} & \textbf{NewsQA} & \textbf{NGQA} \\
% \midrule
% \multicolumn{4}{l}{\textbf{Avg. \# of Nodes:}} \\
% \quad Query   & 1    & 1    & 1    \\
% \quad Agent   & 24   & 24   & 24   \\
% \quad Entity  & 142.1 & 60.3 & 25.0 \\
% \midrule
% \multicolumn{4}{l}{\textbf{Avg. \# of Edges:}} \\
% \quad Entity-Entity  & 385.1 & 109.4 & 26.75  \\
% \quad Agent-Entity   & 123.7 & 142.7 & 123.8 \\
% \quad Query-Entity   & 48.53  & 6.97   & 3.85   \\
% \bottomrule
% \end{tabular}}
% \end{table}
\noindent\textbf{Training signals.}
For each training query, all agents are evaluated and their token-level F1 scores are converted into a soft target distribution via temperature-scaled softmax:
\begin{equation}
  p^*(a \mid q)
  = \frac{\exp\!\bigl(\mathrm{F1}_a(q) / \tau\bigr)}
         {\sum_{a' \in \mathcal{A}}
          \exp\!\bigl(\mathrm{F1}_{a'}(q) / \tau\bigr)}\,.
  \label{eq:soft-target}
\end{equation}
Training minimizes the KL divergence between this target and the router output:
\begin{equation}
  \mathcal{L}_{\mathrm{KL}}(q)
  = \sum_{a \in \mathcal{A}} p^*(a \mid q)\,
    \log \frac{p^*(a \mid q)}{p_\theta(a \mid q, \mathcal{G})}\,.
  \label{eq:kl}
\end{equation}
We use KL divergence because the supervision target is a soft distribution over agents rather than a single best agent. At test time, the router outputs $p_\theta(a \mid q, \mathcal{G})$, and final predictions are assembled by weighted majority voting.
\subsection{Joint Optimization Objective}
\label{subsec:joint}
The formulation above optimizes only the router parameters $\theta$ over a fixed agent pool. We extend it by treating the prompt set
$\boldsymbol{\pi} = \{\pi_a\}_{a \in \mathcal{A}}$ as a jointly optimizable
variable and making the agent budget $K(q)$ query-adaptive at inference time.

Because the soft target $p^*$ depends on the prompts through agent performance, improving prompts reshapes what the router learns and vice versa. The ideal objective jointly optimizes:
\begin{align}
  &\min_{\theta,\, \boldsymbol{\pi}} \;
  \mathbb{E}_{q \sim \mathcal{D}}\!\left[\,
    \mathrm{KL}\!\Big(
      p^*\!\left(\cdot \mid q;\, \boldsymbol{\pi}\right)
      \;\Big\|\;
      p_\theta\!\left(\cdot \mid q, \mathcal{G}\right)
    \Big)
  \,\right],
  \label{eq:joint}
\end{align}
where $p^*(a \mid q;\, \boldsymbol{\pi})$ follows the same softmax form as
Eq.~\ref{eq:soft-target} but now depends on $\boldsymbol{\pi}$ through the agent outputs $\mathrm{F1}_a(q;\, \pi_a)$. Direct optimization of
Eq.~\ref{eq:joint} is intractable: evaluating $\mathrm{F1}_a(q;\, \pi_a)$
requires black-box LLM inference, so no gradient flows from
$\mathcal{L}_{\mathrm{KL}}$ through $\boldsymbol{\pi}$; and prompts live in a
discrete space that precludes continuous relaxation. We therefore alternate
between two sub-problems: (1)~fix $\boldsymbol{\pi}$, train $\theta$ while
collecting diagnostic signals about agent failures (Section~\ref{subsec:diag}); (2)~fix
$\theta$, use these diagnostics to refine the weakest prompts, then retrain
$\theta$ on the updated targets (Section~\ref{subsec:closeloop}). This loop repeats for $R$
rounds. At inference time, the agent budget $K(q)$ is determined adaptively via
an answer-agreement stopping rule (Section~\ref{subsec:adaptive}).

\begin{table*}[t]
\centering
\small
\label{tab:cross-dimension}
\resizebox{0.9\textwidth}{!}{
  \begin{tabular}{@{}lccccc@{}}
  \toprule
  \textbf{Method} & \textbf{System} & \textbf{Routing} & \textbf{Prompt Opt.} &
  \textbf{Training} & \textbf{Optimization} \\
  \midrule
  \rowcolor{gray!20}
\multicolumn{6}{c}{\textit{Adaptive Routing}} \\
LLM-Blender~\citep{jiang-etal-2023-llm}  & Multi-agent  & \cmark & \xmark &
  \cmark & Supervised\\
RouterDC~\citep{10.5555/3737916.3740036}             & Single-agent & \cmark & \xmark &
  \cmark & Supervised \\
  HybridLLM~\citep{ding2024hybridllmcostefficientqualityaware}      & Single-agent & \cmark & \xmark &
  \cmark & Supervised\\
  RouteLLM~\citep{ong2025routellmlearningroutellms}         & Single-agent & \cmark & \xmark &
  \cmark & Supervised \\
  
  MasRouter~\citep{yue2025masrouterlearningroutellms}       & Multi-agent  & \cmark & \xmark &
  \cmark & RL \\
  Router-R1~\citep{zhang2025routerr1teachingllmsmultiround}      & Multi-agent  & \cmark & \xmark &
  \cmark & RL \\
  \midrule
  \rowcolor{gray!20}
\multicolumn{6}{c}{\textit{Structure-aware Routing}} \\GraphRouter~\citep{feng2025graphroutergraphbasedrouterllm}         & Single-agent & \cmark & \xmark &
  \cmark & Supervised \\AgentRouter~\citep{zhang2025agentrouterknowledgegraphguidedllmrouter}         & Multi-agent & \cmark & \xmark &
  \cmark & Supervised \\
  \midrule
  \rowcolor{gray!20}
\multicolumn{6}{c}{\textit{Agent Optimization}} \\
  AFlow~\citep{zhang2025aflowautomatingagenticworkflow}                   & Multi-agent  & \xmark & \cmark &
  \xmark & Search \\
  EvoMAC~\citep{hu2024selfevolvingmultiagentcollaborationnetworks}                & Multi-agent  & \xmark & \cmark &
  \xmark & Evolution \\
  MASS~\citep{zhou2026multiagentdesignoptimizingagents}                 & Multi-agent  & \xmark & \cmark &
  \xmark & Search \\
  MAPRO~\citep{zhang2025maprorecastingmultiagentprompt}
  & Multi-agent  & \xmark & \cmark &
  \xmark & Inference \\
  HiveMind~\citep{xia2025hivemindcontributionguidedonlineprompt}          & Multi-agent  & \xmark & \cmark &
  \xmark & Online \\
  \midrule
  \rowcolor{gray!20}
\multicolumn{6}{c}{\textit{System Design}} \\
  OPTIMA~\citep{chen2025optimaoptimizingeffectivenessefficiency}                 & Multi-agent  & \xmark & \xmark &
  \cmark & Supervised \\
  Multi-Agent Evolve~\citep{chen2025multiagentevolvellmselfimprove}        & Multi-agent  & \xmark & \xmark &
  \cmark & RL \\
  MoA~\citep{wang2024mixtureofagentsenhanceslargelanguage}                   & Multi-agent  & \xmark & \xmark &
  \xmark & Inference \\
  Self-MoA~\citep{li2025rethinkingmixtureofagentsmixingdifferent}            & Single-agent & \xmark & \xmark &
  \xmark & Inference \\
  ADAS~\citep{hu2025automateddesignagenticsystems}                    & Multi-agent  & \xmark & \xmark &
  \xmark & Search \\
  \midrule
  \rowcolor[RGB]{222,230,241}
  \textbf{EvolveRouter (Ours)}               & \textbf{Multi-agent} & \cmark &
  \cmark & \cmark & \textbf{Supervised} \\
  \bottomrule
  \end{tabular}}
    \vspace{-5pt}
  \caption{Cross-dimensional comparison of agent system optimization methods. EvolveRouter is the first framework to jointly support routing, prompt optimization, and training in a multi-agent setting.}
  \vspace{-20pt}
\end{table*}

\begin{figure*}[t]
    \centering
    \includegraphics[width=\textwidth]{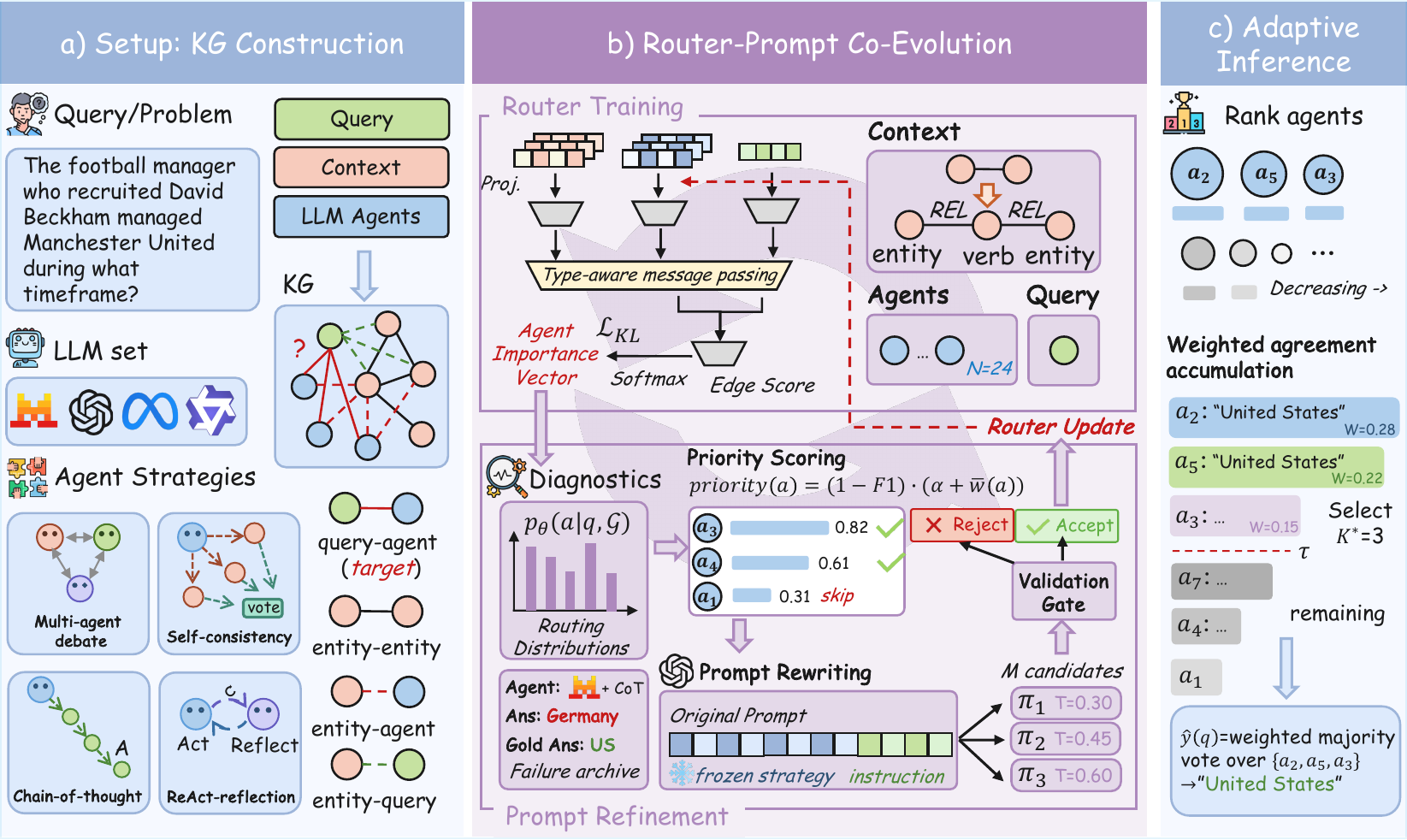}
    \vspace{-15pt}
    \caption{\textbf{Overview of \textsc{EvolveRouter}.} (a) QA instances are converted into knowledge graphs with trainable query–agent edges. (b) RouterGNN learns routing distributions while collecting diagnostics that guide prompt rewriting in a closed-loop co-evolution. (c) Agents are queried by router rank until weighted agreement exceeds $\tau$.}
    \label{fig:framework}
    \vspace{-15pt}
\end{figure*}

\section{Methodology}
\label{sec:method}
We now detail the three stages of the alternating optimization outlined in Section~\ref{subsec:joint}. Figure~\ref{fig:framework} illustrates the overall pipeline.

\subsection{Router Training and Diagnostics}
\label{subsec:diag}
\noindent\textbf{Knowledge graph construction.}
For each QA instance $(q, \mathcal{C})$, the knowledge graph $\mathcal{G}$ defined in Section~\ref{subsec:setup} is instantiated by extracting entity nodes from the context via named entity recognition and dependency parsing. Agent--entity edges are constructed by prompting each agent to identify the entities it attends to. Full construction details in Appendix~\ref{app:impl}.

\noindent\textbf{RouterGNN.}
We instantiate the scoring function $f_\theta$ as a
heterogeneous graph neural network that performs type-aware message passing over
$\mathcal{G}$. Each node embedding is first projected into a shared latent space
through a type-specific projection. For an edge
$(u \xrightarrow{\psi} v)$ of relation type $\psi$, the propagated message is $m_{u \to v}^{(l,\psi)}= \mathrm{Proj}\!\left(W_\psi^{(l)}\, h_u^{(l-1)}\right)$,
and messages are mean aggregated by relation type. Contributions from different edge
types are merged with learnable gate scalars $w_\psi^{(l)}$:
\vspace{-1em}
\begin{equation}
  h_v^{(l)}
  = U_{\tau(v)}^{(l)}\!\left(
      h_v^{(l-1)} \;\Big\|\;
      \sum_{\psi \in \Psi(v)} w_\psi^{(l)} \cdot \bar{m}_v^{(l,\psi)}
    \right),
\end{equation}
where $\tau(v)$ denotes the node type, $U_{\tau(v)}^{(l)}$ is a type-specific
update function, and $\|$ denotes concatenation. After $L$ layers, routing
scores are computed from the final query and agent embeddings: $s(q, a)= \mathrm{MLP}\!\bigl(h_q^{(L)} \| h_a^{(L)}\bigr)$, yielding the distribution $p_\theta(a \mid q, \mathcal{G})$.

\noindent\textbf{Training and diagnostics.}
For each training query, all agents are executed, their answers are scored
against the gold answer, and the resulting F1 values are converted into soft
supervision targets for KL-based router training (Eq.~\ref{eq:kl}). In addition
to learning router parameters, this stage collects diagnostic information for
later refinement.

We collect four types of diagnostic signals during training: agent-level performance summaries per backbone–role pair, a failure archive of incorrect examples with gold answers and router weights, router weight statistics indicating which agents are most relied upon, and a prompt-aware answer cache that automatically invalidates entries when prompts change. These signals guide the refinement stage described next.
\subsection{Closed-Loop Prompt Refinement}
\label{subsec:closeloop}
The goal of prompt refinement is to improve agents whose behavior appears limited by their prompt rather than their backbone capacity.

\noindent\textbf{Target Selection.}
We first apply a role-level consistency filter to distinguish prompt-level weaknesses from backbone-specific outliers. A role becomes eligible for prompt refinement only if it underperforms on multiple backbones. The surviving roles are then expanded to backbone-role pairs and ranked at the agent level. For each backbone-role pair $a$, we compute a refinement priority score: $\mathrm{priority}(a)=\mathrm{severity}(a)\,\bigl(\alpha + \overline{w}(a)\bigr)$, where $\mathrm{severity}(a) = 1 - \overline{\mathrm{F1}}(a)$, and $\overline{w}(a)$ denotes the router's average learned weight for agent $a$, computed from aggregated routing statistics over the training diagnostics. Here $\alpha > 0$ is a small base coefficient that prevents the priority from vanishing when router weight is low.

This rule prioritizes agents that perform poorly but still receive high routing weight. We rank surviving backbone-role pairs by this score and select only the top few candidates in each round, skipping agents that have already frozen after consecutive failed rewrite attempts.

\noindent\textbf{Prompt Rewriting.}
For each selected role, we prompt a rewriter LLM with the agent's current prompt $\pi_a$ and the associated failure evidence from the diagnostic archive. Rewriting is constrained to preserve the role's core reasoning strategy and output format, and may only add, rephrase, or remove a small number of sentences. This keeps the revised prompt close to the original role while allowing targeted correction of failure patterns.

Because prompt rewriting is noisy, each target role is assigned multiple candidate rewrites. Let $\Pi_a = \{\pi_a^{(1)}, \dots, \pi_a^{(M)}\}$ denote the candidate set for role $a$. Candidates are generated at increasing temperatures to balance conservative and exploratory rewrites.

\noindent\textbf{Validation Gate.}
We evaluate candidate prompts on a targeted validation subset $\mathcal{D}_a^{\mathrm{val}}$ centered on the diagnosed failure cases for agent $a$ and supplemented with additional validation samples, and select
    $\pi_a^\star =
\arg\max_{\pi \in \Pi_a}
\mathrm{Perf}(\pi; \mathcal{D}_a^{\mathrm{val}})$,
where $\mathrm{Perf}(\cdot)$ denotes validation performance on this subset. A candidate is accepted only if it does not cause significant regression on the targeted validation subset and yields a non-negative net improvement at the sample level. This conservative rule prevents harmful rewrites while still allowing prompt updates that improve the target agent in aggregate.

\noindent\textbf{Router Update.}
After prompt updates are accepted, the router is fine-tuned on the revised agent pool rather than retrained from scratch. The first refinement round is initialized from the Stage~1 baseline router, and each later round resumes from the previous round's checkpoint. This preserves routing knowledge accumulated in earlier rounds while allowing the model to adapt to modified agents.
The prompt-aware cache makes this update efficient: outputs for unchanged agents are reused, and only modified agents require fresh LLM calls. After fine-tuning, we apply a validation-based safety check and revert the update if the new router causes regression.

% Together, the priority-driven target selection and soft-target retraining form a closed loop: the router's weights guide which prompts to refine, and prompt refinement in turn reshapes the router's weight distribution.

\subsection{Adaptive Inference via Answer Agreement}
\label{subsec:adaptive}
After closed-loop refinement, we use the learned router for efficient inference. The router is reliable for ranking agents, but its output probabilities are less informative when used directly to determine how many agents should be queried. We therefore use answer agreement, rather than probability shape alone, as the stopping signal.

Let $\sigma$ denote the ranking of agents induced by the router, and let $p_{\sigma(i)}(q) \triangleq p_\theta(a_{\sigma(i)} \mid q, \mathcal{G})$ be the router weight assigned to the i-th queried agent. We query agents sequentially in this order and define the agreement score after consulting the top $k$ agents as
\begin{equation}
    A(k, q)
=
\frac{
\max_{y}
\sum_{i=1}^{k} p_{\sigma(i)}(q)\,\mathbf{1}\!\left[y_{\sigma(i)}(q)=y\right]
}{
\sum_{i=1}^{k} p_{\sigma(i)}(q)
},
\end{equation}
\vspace{-5pt}

where $\mathbf{1}[\cdot]$ is the indicator function. This score measures the weighted share of the most supported answer among the queried agents.

We stop as soon as the agreement score exceeds a threshold $\tau$ after at least $k_{\min}$ agents have been queried:
\begin{equation}
    k^\star(q)
=
\min
\left\{
k \geq k_{\min} : A(k, q) \geq \tau
\right\},
\qquad \text{subject to } k \leq k_{\max}.
\end{equation}
The final prediction is formed by router-weighted aggregation over the consulted agents only. If no early stopping condition is met by $k_{\max}$, we simply use the full consulted set. The hyperparameters $\tau$, $k_{\min}$, and $k_{\max}$ are tuned on the validation split for each dataset.

The router therefore contributes in two ways at inference time: it determines which agents are consulted first, and its learned weights define how strongly each queried answer contributes to agreement. In easy cases, top-ranked agents quickly concentrate weighted support on the same answer and inference terminates early. In harder cases, support remains diffuse and the method allocates a larger budget.

\section{Experiments}
\subsection{Experiment Setup}
\noindent\textbf{Benchmarks.}
We evaluate on five question answering benchmarks with different reasoning demands:
2WikiMultihopQA~\citep{ho2020constructingmultihopqadataset} and HotpotQA~\citep{yang2018hotpotqadatasetdiverseexplainable} require multi-hop reasoning across multiple pieces of evidence.
NewsQA~\citep{trischler2017newsqamachinecomprehensiondataset} and TriviaQA~\citep{joshi2017triviaqalargescaledistantly} emphasize factual comprehension and retrieval-oriented answering.
NGQA~\citep{zhang2024ngqanutritionalgraphquestion} provides an additional domain-specific setting with answer patterns that differ from the other four benchmarks. We describe the details of these benchmarks and discuss the split of sets with other details in Appendix~\ref{app:benchmarks}.

\noindent\textbf{Agent Pool.}
We build the agent pool by pairing four backbone LLMs with six prompting roles.
The four backbones are Llama-3-8B-Instruct~\citep{meta2024llama3}, Qwen2.5-7B-Instruct-Turbo~\citep{qwen2025qwen25}, Mixtral-8$\times$7B-Instruct~\citep{mistralai2023mixtral}, and gpt-oss-20b~\citep{openai2025gptoss20b}.
The six roles are Raw (the basic LLM method), Chain-of-Thought (CoT)~\citep{wei2022chain}, Self-Consistency (SC)~\citep{wang2023selfconsistencyimproveschainthought}, Multi-Agent Debate (MAD)~\citep{du2024improving}, React-Reflection~\citep{yao2022react,shinn2023reflexion}, and Multi-Agent Summary.

\noindent\textbf{Evaluation Metrics.}
Following standard practice on SQuAD~\citep{rajpurkar2016squad100000questionsmachine}, we report Exact Match (EM) and token-level F1 on all benchmarks.
EM measures whether the normalized prediction exactly matches the gold answer.
F1 gives partial credit when the predicted answer overlaps with the gold answer but is not an exact match.

\noindent\textbf{Baselines.} We compare against four groups: (1)~Heuristic baselines (Average, Majority Vote, Best LLM, Best Agent); (2)~Training-free Routing (RandomRouter, CascadeRouting~\citep{dekoninck2025unifiedapproachroutingcascading}, RouteLLM~\citep{ong2025routellmlearningroutellms}); (3)~Adaptive Routing (LLM-Blender~\citep{jiang-etal-2023-llm}, HybridLLM~\citep{ding2024hybridllmcostefficientqualityaware}, KNN Router~\citep{hu2024routerbenchbenchmarkmultillmrouting}, MLP Router~\citep{hu2024routerbenchbenchmarkmultillmrouting}); and (4)~Structure-aware routing (GraphRouter~\citep{feng2025graphroutergraphbasedrouterllm}, AgentRouter~\citep{zhang2025agentrouterknowledgegraphguidedllmrouter}). We also report an \textbf{Oracle} upper bound. Details of each baseline are provided in Appendix~\ref{app:baselines}.
\subsection{Main Results}
We present the main results in Table~\ref{tab:main_results}. Across all five benchmarks, EvolveRouter consistently achieves the strongest performance among all practical baselines, improving over the best structure-aware routing baseline on every dataset in both F1 and EM. The largest gain appears on TriviaQA, where EvolveRouter improves over AgentRouter by +7.08 F1 and +8.67 EM. This suggests that TriviaQA is particularly sensitive to prompt-task alignment and that better prompts can substantially lift the quality of routed candidates. Among routing baselines, flat methods (KNN Router, MLP Router, RouteLLM) improve over heuristic aggregation but still fall behind EvolveRouter on most benchmarks. This indicates that modeling structured query-entity-agent relations provides advantages that shallower strategies cannot replicate.

EvolveRouter also outperforms the Best Agent baseline on all five benchmarks. Since Best Agent already represents the strongest single agent in the pool, this confirms that routed collaboration still contributes complementary information even after prompts are refined. Prompt optimization and multi-agent routing are not substitutes but reinforcing components. Finally, EvolveRouter exhibits lower variance than AgentRouter across most datasets. For example, the F1 standard deviation on TriviaQA drops from 3.69 to 0.32, indicating that prompt refinement improves not only average performance but also system consistency. The remaining gap to the Oracle suggests significant headroom for future work.

\begin{table*}[t]
\centering
% \scriptsize
% \setlength{\tabcolsep}{2.5pt}
% \renewcommand{\arraystretch}{1.12}

% % thickness configs
% \newcommand{\thickhline}{\specialrule{1.2pt}{0pt}{0pt}} 
% \newcommand{\midsep}{\specialrule{0.6pt}{0pt}{0pt}}    
% \newcommand{\doublehline}{%
%   \specialrule{1.2pt}{0pt}{0pt}%
%   \specialrule{1.2pt}{1.2pt}{0pt}%
% }
\resizebox{\textwidth}{!}{
\begin{tabular}{l cc cc cc cc cc}
\toprule
\multirow{2}{*}{\textbf{Method}} &
\multicolumn{2}{c}{\textbf{2Wiki}} &
\multicolumn{2}{c}{\textbf{HotpotQA}} &
\multicolumn{2}{c}{\textbf{NewsQA}} &
\multicolumn{2}{c}{\textbf{TriviaQA}} &\multicolumn{2}{c}{\textbf{NGQA}} \\
\cmidrule(lr){2-3}\cmidrule(lr){4-5}\cmidrule(lr){6-7}\cmidrule(lr){8-9}\cmidrule(lr){10-11}
& \textbf{F1} & \textbf{EM}
& \textbf{F1} & \textbf{EM}
& \textbf{F1} & \textbf{EM}
& \textbf{F1} & \textbf{EM} 
& \textbf{F1} & \textbf{EM}\\
\toprule
\rowcolor{gray!20}
\multicolumn{11}{c}{\textit{Heuristic Baselines}} \\
Average
&52.15{\tiny$\pm$2.20} & 42.73{\tiny$\pm$1.88}
&59.52{\tiny$\pm$1.09} & 46.63{\tiny$\pm$0.79}
&57.97{\tiny$\pm$1.64} & 36.10{\tiny$\pm$0.69}
&45.69{\tiny$\pm$2.27}& 37.32{\tiny$\pm$1.64}
&47.50{\tiny$\pm$0.80}&4.04{\tiny$\pm$0.38}
\\
Majority Vote
&67.67{\tiny$\pm$0.52} &58.55{\tiny$\pm$0.55} 
&63.88{\tiny$\pm$1.24} &52.89{\tiny$\pm$0.92}
&59.30{\tiny$\pm$0.64} &39.50{\tiny$\pm$1.29} 
&53.73{\tiny$\pm$0.72} &43.00{\tiny$\pm$0.80} 
&48.42{\tiny$\pm$0.38} &4.00{\tiny$\pm$0.44} 
\\
Best LLM
 & 71.77{\tiny$\pm$1.71}&62.72{\tiny$\pm$1.32}
 &68.19{\tiny$\pm$1.37} &51.79{\tiny$\pm$2.09}
&60.55{\tiny$\pm$1.45} &39.80{\tiny$\pm$2.05} 
&51.41{\tiny$\pm$0.65}&42.09{\tiny$\pm$0.59}
&48.03{\tiny$\pm$0.74} &3.17{\tiny$\pm$0.58}
 \\
Best Agent
 &75.21{\tiny$\pm$1.11} &64.56{\tiny$\pm$1.91}
 &68.74{\tiny$\pm$1.39} &56.15{\tiny$\pm$0.87}
&62.80{\tiny$\pm$2.15} &37.11{\tiny$\pm$1.15} 
&60.27{\tiny$\pm$2.52}
&50.13{\tiny$\pm$2.01}
&48.41{\tiny$\pm$0.71}&4.00{\tiny$\pm$1.00}\\

\rowcolor{gray!20}
\multicolumn{11}{c}{\textit{Training-free Routing}} \\
Random Router
&54.76{\tiny$\pm$0.37}&45.00{\tiny$\pm$0.58}
&60.59{\tiny$\pm$1.88}&47.90{\tiny$\pm$1.15}
&58.11{\tiny$\pm$0.95}&36.20{\tiny$\pm$0.73}
&46.05{\tiny$\pm$2.45}&37.00{\tiny$\pm$1.00}
&47.41{\tiny$\pm$0.27}&4.03{\tiny$\pm$0.12}
\\
CascadeRouting
&73.36{\tiny$\pm$1.03}&65.33{\tiny$\pm$1.53}
&65.57{\tiny$\pm$0.20}&52.67{\tiny$\pm$0.58}
&60.17{\tiny$\pm$0.32}&39.00{\tiny$\pm$0.88}
&61.86{\tiny$\pm$1.59}&50.33{\tiny$\pm$1.53}
&47.06{\tiny$\pm$0.42}&3.33{\tiny$\pm$0.58}
\\
RouteLLM
&71.49{\tiny$\pm$1.36}&65.00{\tiny$\pm$1.00}
&68.27{\tiny$\pm$0.16}&55.33{\tiny$\pm$0.58}
&59.93{\tiny$\pm$0.34}&38.00{\tiny$\pm$0.00}
&62.54{\tiny$\pm$0.16}&51.33{\tiny$\pm$0.58}
&47.20{\tiny$\pm$0.93}&4.00{\tiny$\pm$1.00}
\\

\rowcolor{gray!20}
\multicolumn{11}{c}{\textit{Adaptive Routing}} \\

LLM-Blender
&64.03{\tiny$\pm$0.55} & 49.67{\tiny$\pm$0.58}
&64.22{\tiny$\pm$2.21} & 49.00{\tiny$\pm$1.73}
&59.30{\tiny$\pm$1.09} &37.33{\tiny$\pm$2.89} 
&52.75{\tiny$\pm$0.94}&39.33{\tiny$\pm$1.53}
&42.15{\tiny$\pm$1.07}&1.67{\tiny$\pm$0.47}
 \\
HybridLLM
&66.91{\tiny$\pm$0.42} & 54.67{\tiny$\pm$0.58}
&60.87{\tiny$\pm$0.39} & 42.67{\tiny$\pm$0.58}
&54.19{\tiny$\pm$0.49} &31.00{\tiny$\pm$1.00} 
&55.62{\tiny$\pm$0.71} & 43.00{\tiny$\pm$1.00}
&48.05{\tiny$\pm$0.36}&2.33{\tiny$\pm$0.47}
\\
KNN Router
&71.54{\tiny$\pm$0.13}&63.00{\tiny$\pm$0.00}
&64.72{\tiny$\pm$0.15}&52.00{\tiny$\pm$0.00}
&57.15{\tiny$\pm$0.40}&34.00{\tiny$\pm$0.00}
&58.02{\tiny$\pm$0.44}&48.00{\tiny$\pm$0.00}
&46.11{\tiny$\pm$0.77}&4.00{\tiny$\pm$1.00}
\\
MLP Router
&72.59{\tiny$\pm$0.07}&64.00{\tiny$\pm$0.00}
&68.91{\tiny$\pm$0.45}&56.33{\tiny$\pm$0.58}
&60.94{\tiny$\pm$0.18}&37.00{\tiny$\pm$0.00}
&48.28{\tiny$\pm$0.31}&39.33{\tiny$\pm$0.58}
&48.80{\tiny$\pm$0.56}&4.33{\tiny$\pm$0.58}
\\
\rowcolor{gray!20}
\multicolumn{11}{c}{\textit{Structure-aware Routing}} \\

GraphRouter
&61.98{\tiny$\pm$0.42} & 53.00{\tiny$\pm$1.00}
&60.87{\tiny$\pm$1.82} & 48.33{\tiny$\pm$1.52}
&62.66{\tiny$\pm$2.59} &42.33{\tiny$\pm$0.58}
&50.95{\tiny$\pm$3.01} & 41.67{\tiny$\pm$1.53}
&46.88{\tiny$\pm$0.42}& 4.17{\tiny$\pm$0.21}\\

AgentRouter
&75.54{\tiny$\pm$0.44}&67.67{\tiny$\pm$0.58}
&69.35{\tiny$\pm$0.56}&56.33{\tiny$\pm$0.58}
&63.02{\tiny$\pm$2.44}&41.33{\tiny$\pm$0.58}
&62.75{\tiny$\pm$3.69}&51.33{\tiny$\pm$2.51}
&49.95{\tiny$\pm$1.29}&8.67{\tiny$\pm$1.53}
\\
% AgentRouter
% & 72.54$\pm$0.50&64.50$\pm$0.56
% &70.50$\pm$0.20&57.25$\pm$0.50
% &64.52$\pm$2.44&46.50$\pm$0.45
% &63.77$\pm$0.83&52.00$\pm$2.13
% \\
% K=3
% &75.54$\pm$0.44&67.75$\pm$1.26
% &68.08$\pm$1.07&55.00$\pm$1.42
% &66.39$\pm$0.37&49.50$\pm$0.56
% &60.33$\pm$1.54&50.00$\pm$2.70
% \\
% K=5
% &77.01$\pm$0.47&69.75$\pm$1.89
% &69.35$\pm$0.56&56.50$\pm$0.58
% &65.91$\pm$0.44&49.00$\pm$0.53
% &59.95$\pm$1.86&48.75$\pm$2.32
% \\
% K=10
% &75.61$\pm$1.74&69.00$\pm$3.46
% &70.15$\pm$0.50&56.75$\pm$0.50
% &64.53$\pm$0.40&47.25$\pm$0.55
% &62.75$\pm$3.69&51.25$\pm$4.92
% \\
\midrule
\rowcolor[RGB]{222,230,241}
\textbf{EvolveRouter}
&\textbf{77.70{\tiny$\pm$0.13}}&\textbf{70.33{\tiny$\pm$0.58}}
&\textbf{72.28{\tiny$\pm$0.86}}&\textbf{60.33{\tiny$\pm$1.25}}
&\textbf{64.74{\tiny$\pm$0.77}}&41.67{\tiny$\pm$0.47}
&\textbf{69.83{\tiny$\pm$0.32}}&\textbf{60.00{\tiny$\pm$1.00}}
&\textbf{54.04{\tiny$\pm$0.60}}&\textbf{9.67{\tiny$\pm$0.58}}
\\
w/o PO
&\underline{76.69{\tiny$\pm$0.31}}&\underline{68.00{\tiny$\pm$0.00}}
&70.41{\tiny$\pm$0.50}&57.00{\tiny$\pm$0.58}
&63.81{\tiny$\pm$0.83}&41.33{\tiny$\pm$0.58}
&63.10{\tiny$\pm$0.51}&54.67{\tiny$\pm$0.58}
&50.30{\tiny$\pm$1.74}&9.00{\tiny$\pm$1.00}
\\
w/o Adaptive K
&76.53{\tiny$\pm$0.19}&67.67{\tiny$\pm$0.58}
&\underline{71.95{\tiny$\pm$1.38}}&\underline{58.33{\tiny$\pm$1.15}}
&\underline{64.48{\tiny$\pm$0.52}} &\textbf{42.00{\tiny$\pm$0.00}}
&\underline{68.53{\tiny$\pm$1.53}}&\underline{58.67{\tiny$\pm$1.53}}
&\underline{53.28{\tiny$\pm$0.16}}&\underline{9.67{\tiny$\pm$0.58}}
\\
\midrule
Oracle
&91.79{\tiny$\pm$2.58}& 86.00{\tiny$\pm$2.65}
&87.80{\tiny$\pm$0.59} &79.67{\tiny$\pm$0.58} 
&80.50{\tiny$\pm$1.75} &60.67{\tiny$\pm$1.53}
&72.77{\tiny$\pm$0.71} &60.67{\tiny$\pm$1.03}
&69.29{\tiny$\pm$2.35}&27.00{\tiny$\pm$1.00}\\
\bottomrule
% temp=0.9 + OpenRouter
% \\
% AgentRouter 
% &74.65 & 66.00
% &66.85 & 57.00
% &61.38 &46.00 
% & & \\
% K=1
% &72.14 & 63.00
% &66.28 & 54.00
% &59.76 & 43.00
% & & \\
% K=3
%  &73.78 &65.00
% &67.00 & 55.00
% &57.52 &39.00 
% & & \\
% K=5
%  &73.65 &65.00
%  &66.18 &56.00
% & 57.52&40.00 
% & & \\
% K=10
%  &74.65 &66.00
%  &65.85 &56.00
% &59.45 &43.00 
% && \\
% +PO
%  &73.50&64.00
% &67.00&55.00\\
% +Adaptive K
% &73.50&64.00
% &66.19&56.00\\
% Oracle
% &86.03 &81.00 
% &77.58 &69.00 
% &79.59 & 65.00
% & & \\
% \thickhline

% temp=0.6 + TogetherAI\\
% AgentRouter
% &75.16 &66.00\\
% K=1
% &72.69&64.00\\
% K=3
% &75.49&66.00
% \\
% K=5
% &77.16&68.00
% \\
% K=10
% &76.16&67.00
% \\
% Oracle
% &84.29 &79.00 
% & & 
% & &
% & & \\
% \thickhline

\end{tabular}
}
\vspace{-5pt}
\caption{Performance results with baseline methods on the five QA benchmarks. We report the mean and standard deviation for all results. Best results (excluding Oracle) are in \textbf{bold}, second best are \underline{underlined}.}
\label{tab:main_results}
\vspace{-15pt}
\end{table*}

\subsection{Ablation Studies}
Table~\ref{tab:ablation_po} studies the contribution of each component. Starting from the KG Router base, each round of PO improves performance on most datasets, with cumulative gains reaching +1.31\% F1 on 2Wiki, +3.75\% on HotpotQA, +9.21\% on TriviaQA, and +6.67\% on NGQA. The largest improvements appear on TriviaQA and NGQA, suggesting these benchmarks are particularly sensitive to prompt-task alignment. The round-by-round results also reveal varying convergence speeds: on 2Wiki the first round already captures the full gain, while HotpotQA, NewsQA, and NGQA continue to benefit from subsequent rounds, indicating that some prompt weaknesses require multiple iterations to resolve.

Adaptive K contributes a consistent but more modest improvement. Compared with the base, adding Adaptive K alone improves F1 by +1.52\% on 2Wiki, +1.53\% on HotpotQA, +1.25\% on NewsQA, and +0.70\% on NGQA. The effect is smaller than PO on every dataset, confirming that the larger share of the final gain comes from improving agent quality rather than the stopping strategy. Nonetheless, the two components are complementary: PO makes the top-ranked agents more reliable, which in turn strengthens the answer agreement signal for Adaptive K. This explains why the full model consistently outperforms either component alone.

\begin{wrapfigure}{r}{0.37\textwidth}
\vspace{-1pt}
\centering
\includegraphics[width=\linewidth]{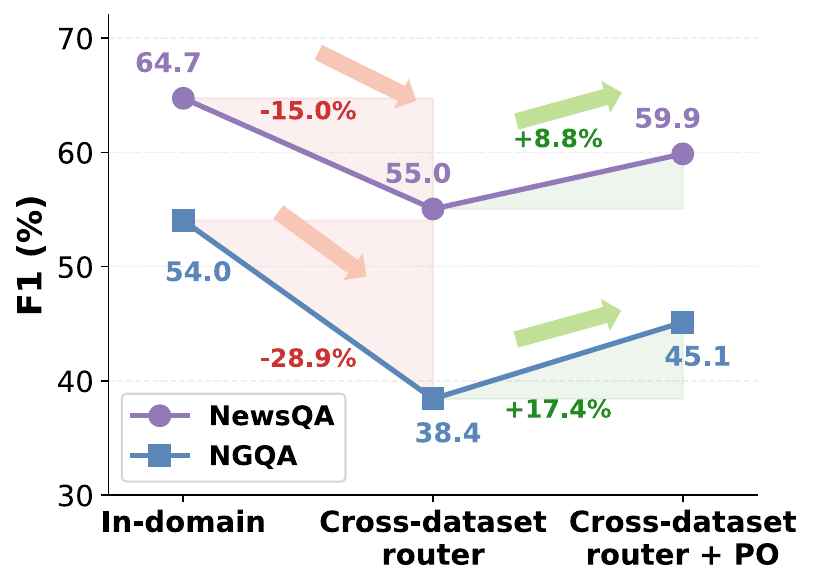}
\vspace{-10pt}
\caption{Cross-dataset transfer of the router trained on HotpotQA.}
\label{fig:transfer}
% \vspace{-10pt}
\end{wrapfigure}

As shown in Table~\ref{tab:ablation_po} and Figure~\ref{fig:adaptive_k}, no single global $K$ performs best across all datasets, and Adaptive K addresses this by determining $K$ per query through answer agreement, exceeding the best fixed-$K$ on every benchmark. We further examine transferability in Figure~\ref{fig:transfer}: routing knowledge does not transfer well across benchmarks, though incorporating optimized prompts from the source task can partially mitigate the degradation. This suggests that both routing and prompt quality are task-dependent and benefit from joint per-dataset optimization. Further details are provided in Appendix~\ref{app:additional}.

% \begin{figure}[t]
% \centering
% \includegraphics[width=0.44\textwidth]{transfer.pdf}
% \caption{Cross-dataset transferability of the router trained on HotpotQA.}
% \label{fig:transfer}
% \end{figure}

\begin{table*}[t]
\centering
\resizebox{\textwidth}{!}{
\begin{tabular}{l cc cc cc cc cc}
\toprule
\multirow{2}{*}{\textbf{Method}} &
\multicolumn{2}{c}{\textbf{2Wiki}} &
\multicolumn{2}{c}{\textbf{HotpotQA}} &
\multicolumn{2}{c}{\textbf{NewsQA}} &
\multicolumn{2}{c}{\textbf{TriviaQA}} &
\multicolumn{2}{c}{\textbf{NGQA}} \\
\cmidrule(lr){2-3}\cmidrule(lr){4-5}\cmidrule(lr){6-7}\cmidrule(lr){8-9}\cmidrule(lr){10-11}
& \textbf{F1} & \textbf{EM}
& \textbf{F1} & \textbf{EM}
& \textbf{F1} & \textbf{EM}
& \textbf{F1} & \textbf{EM}
& \textbf{F1} & \textbf{EM} \\
\toprule
% STRATCH (all backbone)
% \\
% w/o PO
% &76.45  &67.00 
% &69.36  &56.00 
% &59.80  &40.00 
% &59.47  &49.00 
% & -- & -- \\
% +PO (1st round)
% &75.78  &67.00 
% &70.13  &57.00 
% & 61.96 &39.00 
% &70.07  &60.00 
% & -- & -- \\
% +PO (2nd round)
% &75.78  & 67.00
% & 70.52 & 58.00
% & 62.26 & 40.00
% & 69.23 & 59.00
% & -- & -- \\
% +PO (3rd round)
% & 76.70 & 68.00
% & 69.40 & 57.00
% & 63.15 & 42.00
% & -- & --
% & -- & -- \\
% \thickhline
KG Router (base)
&75.54{\tiny$\pm$0.44}&67.67{\tiny$\pm$0.58}
&69.35{\tiny$\pm$0.56}&56.33{\tiny$\pm$0.58}
&63.02{\tiny$\pm$2.44}&41.33{\tiny$\pm$0.58}
&62.75{\tiny$\pm$3.69}&51.33{\tiny$\pm$2.51}
&49.95{\tiny$\pm$1.29}&8.67{\tiny$\pm$1.53}
\\
\rowcolor{gray!20}
\multicolumn{11}{c}{\textit{Prompt Optimization Ablation}} \\

+PO (Round 1) 
&76.53{\tiny$\pm$0.19}&67.67{\tiny$\pm$0.58}
&70.31{\tiny$\pm$0.77}&57.33{\tiny$\pm$0.58}
&63.48{\tiny$\pm$0.75}&41.67{\tiny$\pm$0.58}
&67.00{\tiny$\pm$0.43}&57.00{\tiny$\pm$1.00}
&50.31{\tiny$\pm$0.88}&9.00{\tiny$\pm$0.00}

\\
+PO (Round 2)
&76.53{\tiny$\pm$0.19}&67.67{\tiny$\pm$0.58}
% &72.28&60.00
&71.95{\tiny$\pm$1.38}&58.33{\tiny$\pm$1.15}
&64.07{\tiny$\pm$1.02}&42.00{\tiny$\pm$1.00}
&68.53{\tiny$\pm$1.53}&58.67{\tiny$\pm$1.53}
&52.99{\tiny$\pm$1.19}&9.33{\tiny$\pm$0.58}
\\
+PO (Round 3)
&--&--
&--&--
&64.48{\tiny$\pm$0.52} &42.00{\tiny$\pm$0.00}
&--&--
&53.28{\tiny$\pm$0.16}&9.67{\tiny$\pm$0.58}
\\
\rowcolor[RGB]{222,230,241}
\textbf{$\Delta$ vs.\ base (\%)} 
  & $\uparrow$ 1.31 \% & $\uparrow$ 0.00\% & $\uparrow$ 3.75\%&$\uparrow$ 3.55\%&$\uparrow$ 2.32\%&$\uparrow$ 1.62\%&$\uparrow$ 9.21\%&$\uparrow$ 14.30\%& $\uparrow$ 6.67\% &$\uparrow$ 11.53\%\\
\midrule
+PO (from scratch)
&75.78{\tiny$\pm$2.31}  &67.33{\tiny$\pm$0.58}
&70.07{\tiny$\pm$1.52}  &57.33{\tiny$\pm$1.15}
& 63.15{\tiny$\pm$1.74} & 41.00{\tiny$\pm$1.00}
&65.17{\tiny$\pm$0.93}&54.67{\tiny$\pm$1.15}
&48.07{\tiny$\pm$1.32}&8.67{\tiny$\pm$1.53}
\\
+PO (worst agent only)
&76.17{\tiny$\pm$0.92}&67.33{\tiny$\pm$1.15}
&70.41{\tiny$\pm$2.25}&57.00{\tiny$\pm$1.00}
&61.80{\tiny$\pm$0.88}&40.33{\tiny$\pm$0.58}
&59.47{\tiny$\pm$1.55}&49.00{\tiny$\pm$1.00}
&50.03{\tiny$\pm$0.47}&8.67{\tiny$\pm$0.58}
\\
\midrule
\rowcolor{gray!20}
\multicolumn{11}{c}{\textit{Adaptive K Ablation}} \\
+ Adaptive K
&76.69{\tiny$\pm$0.31}&68.00{\tiny$\pm$0.00}
&70.41{\tiny$\pm$0.50}&57.00{\tiny$\pm$0.58}
&63.81{\tiny$\pm$0.83}&41.33{\tiny$\pm$0.58}
&63.10{\tiny$\pm$0.51}&54.67{\tiny$\pm$0.58}
&50.30{\tiny$\pm$1.74}&9.00{\tiny$\pm$1.00}
\\
\rowcolor[RGB]{222,230,241}
\textbf{$\Delta$ vs.\ base (\%)} 
&$\uparrow$ 1.52\%&$\uparrow$ 0.49\%&$\uparrow$ 1.53\%&$\uparrow$ 1.19\%&$\uparrow$ 1.25\%&$\uparrow$ 0.00\%&$\uparrow$ 0.56\%&$\uparrow$ 6.51\%&$\uparrow$ 0.70\%&$\uparrow$ 3.81\%
% \\
% Fixed K=3
% &75.49{\tiny$\pm$0.59}&66.67{\tiny$\pm$0.58}
% &67.08{\tiny$\pm$1.17}&54.00{\tiny$\pm$1.00}
% &60.92{\tiny$\pm$0.68}&40.67{\tiny$\pm$0.58}
% &61.60{\tiny$\pm$0.86}&51.00{\tiny$\pm$1.00}
% &49.76{\tiny$\pm$0.49}&8.67{\tiny$\pm$0.58}
% \\
% Fixed K=5
% &76.22{\tiny$\pm$1.46}&67.00{\tiny$\pm$1.00}
% &69.83{\tiny$\pm$1.36}&56.67{\tiny$\pm$0.58}
% &62.83{\tiny$\pm$0.46}&41.00{\tiny$\pm$1.00}
% &62.60{\tiny$\pm$0.81}&52.33{\tiny$\pm$0.58}
% &48.94{\tiny$\pm$0.59}&7.33{\tiny$\pm$0.58}
% \\
% Fixed K=10
% &76.16{\tiny$\pm$0.81}&67.00{\tiny$\pm$1.00}
% &69.40{\tiny$\pm$0.88}&56.00{\tiny$\pm$1.00}
% &61.08{\tiny$\pm$1.38}&41.00{\tiny$\pm$1.00}
% &61.18{\tiny$\pm$0.76}&49.67{\tiny$\pm$1.15}
% &48.34{\tiny$\pm$0.31}&7.00{\tiny$\pm$0.00}
\\
% FINE-TUNE (one backbone)
% \\
% w/o PO
% &76.45&67.00
% &69.36&56.00
% &59.80&40.00
% \\
% +PO (1st round)
% &68.59&--
% &70.41&57.00
% &60.82&40.00
% \\
% +PO (2nd round)
% &76.17&67.00
% &70.41&57.00
% &61.80&42.00
% \\
% +PO (3rd round)
% &75.17&66.00
% &70.41&57.00
% &61.70&41.00
% \\
\bottomrule
\end{tabular}
}
\vspace{-5pt}
\caption{Ablation study of prompt optimization and adaptive K across five benchmarks.}
\label{tab:ablation_po}
\vspace{0.3em}
\includegraphics[width=\textwidth]{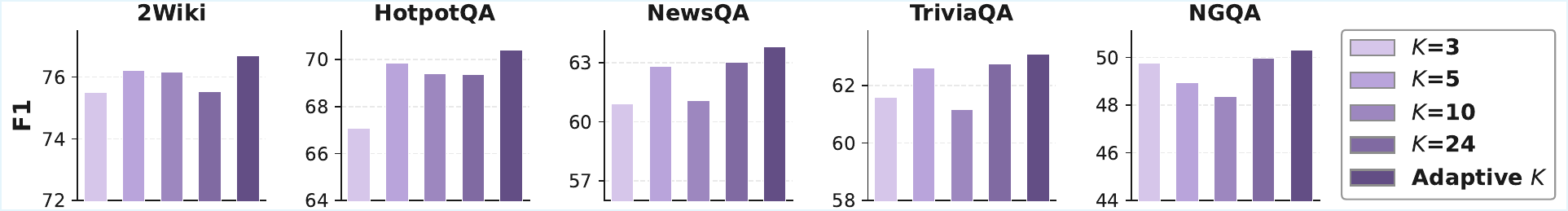}
\vspace{-15pt}
\captionof{figure}{F1 comparison of Adaptive K against fixed-$K$ settings.}
\vspace{-20pt}
\label{fig:adaptive_k}
\end{table*}
\subsection{Analysis}
\textbf{What prompt optimization changes.} Inspecting the accepted rewrites reveals several recurring patterns: forcing a best-guess answer instead of declining, explicit entity extraction and tracking, upgrading vague instructions into numbered step lists, and adding domain-specific format rules. All accepted rewrites modify only the instruction suffix while preserving the role's core strategy structure. Concrete before-and-after examples are provided in Appendix~\ref{app:case}.

\textbf{The worst agent is not the most valuable to fix.} A natural heuristic is to always rewrite the agent with the lowest F1. However, Table~4 shows that this strategy yields marginal gains on 2Wiki and HotpotQA, slightly hurts NewsQA, and barely changes NGQA. Our diagnostic-guided selection improves all five datasets by a clear margin. The key difference is that our priority score accounts not only for how poorly an agent performs, but also for how heavily the router relies on it. An agent with low F1 that the router already avoids has little effect on the final ensemble. Conversely, rewriting an agent with low F1 but high router weight produces the largest marginal gain, because the router amplifies both its improvements and its errors.

\noindent\textbf{Routing knowledge must be preserved across co-evolution rounds.}
Retraining the router from scratch after prompt optimization consistently underperforms checkpoint resuming, with F1 gaps of +0.75 on 2Wiki, +1.88 on HotpotQA, +0.92 on NewsQA, and +4.92 on NGQA (Table~\ref{tab:ablation_po}). This is notable because only 2--3 out of 24 agents are modified per round. Retraining from scratch forces the router to relearn routing patterns for the 21--22 unchanged agents, discarding useful knowledge. Checkpoint resuming preserves these patterns while adapting only to the modified agents. This result suggests that routing and prompt optimization are \emph{tightly coupled}: the router's learned structure provides a stable scaffold that prompt refinement should perturb minimally rather than disrupt.

\noindent\textbf{Computational cost.}
The GNN training takes under 5 minutes on a single GPU, and each refinement round only re-queries the 2--3 modified agents while reusing cached outputs for the rest. At inference time, Adaptive~K reduces the average number of agent calls from 24 to 2.9--7.4 across our benchmarks, yielding a 69--88\% reduction while maintaining or improving F1. Further cost analysis is provided in Appendix~\ref{app:training}.

\section{Conclusion}
We presented EvolveRouter, a framework in which routing and agent quality co-evolve: router diagnostics identify underperforming agents and guide targeted prompt revisions, while improved agents provide cleaner supervision for the next round of router training. An answer-agreement stopping rule further adapts the number of queried agents per query. Experiments on five QA benchmarks show consistent gains over strong routing baselines in both F1 and EM. Our analysis reveals that routing-guided agent selection outperforms naive lowest-F1 heuristics, and that routing knowledge must be preserved across refinement rounds. Future directions include expanding the agent pool dynamically and incorporating automated agent design.

% \section*{Author Contributions}

% \section*{Acknowledgments}
% Use unnumbered first level headings for the acknowledgments. All

% \section*{Ethics Statement}
% Authors can add an optional ethics statement to the paper. 
% For papers that touch on ethical issues, this section will be evaluated as part of the review process. The ethics statement should come at the end of the paper. It does not count toward the page limit, but should not be more than 1 page. 

\bibliography{colm2026_conference}
\bibliographystyle{colm2026_conference}

\appendix
\section{Related Work}
\subsection{LLM Routing and Multi-Agent Collaboration}

Combining outputs from multiple LLMs to outperform any single model has become an active area of research. Early work mainly relied on heuristic aggregation. Mixture-of-Agents~\citep{wang2024mixtureofagentsenhanceslargelanguage} showed that stacking agents in layers with equal weights can outperform leading models. Self-Consistency~\citep{wang2023selfconsistencyimproveschainthought} and LLM-Blender~\citep{jiang-etal-2023-llm} introduced majority voting and pairwise ranking, respectively. However, these methods apply the same aggregation strategy to all queries and therefore lack adaptivity.

This limitation motivated learned routers. RouteLLM~\citep{ong2025routellmlearningroutellms}, RouterDC~\citep{10.5555/3737916.3740036}, and MixLLM~\citep{wang2025mixllmdynamicroutingmixed} learn query-level routing decisions through preference learning, contrastive learning, and contextual bandits. More structure-aware methods further model collaboration explicitly. MasRouter~\citep{yue2025masrouterlearningroutellms} uses a cascaded controller to determine collaboration mode, role assignment, and model selection. GraphRouter~\citep{feng2025graphroutergraphbasedrouterllm} formulates routing as link prediction on a heterogeneous graph, while G-Designer~\citep{zhang2025gdesignerarchitectingmultiagentcommunication} dynamically generates communication topologies for each task. Another line of work treats the router itself as an LLM. Router-R1~\citep{zhang2025routerr1teachingllmsmultiround} and Nielsen et al.~\citep{nielsen2026learningorchestrateagentsnatural} train LLM-based orchestrators with reinforcement learning, but these methods incur substantial inference overhead because each routing decision requires a full LLM forward pass.

Among graph-based approaches \citep{shi2026ng, feng2025graphroutergraphbasedrouterllm}, Zhang et al.~\citep{zhang2025agentrouterknowledgegraphguidedllmrouter} encode queries, entities, and agents into a knowledge graph and train a heterogeneous GNN for per-query routing. While effective, this and all methods above assume a fixed agent pool. Our work starts from the observation that routing effectiveness is ultimately bounded by agent quality, and introduces a closed-loop framework that jointly improves both.

\subsection{Prompt Optimization for Multi-Agent Systems}
The sensitivity of LLMs to prompt wording is well documented. Format changes alone can produce large accuracy swings~\citep{sclar2024quantifyinglanguagemodelssensitivity}. Semantically equivalent rephrasings can degrade the worst-performing prompt to near-random performance~\citep{cao2024worstpromptperformancelarge}. Even a 671B-parameter model trained with reinforcement learning remains highly sensitive to prompt variation~\citep{Guo_2025}. These findings show that prompt quality can be a major determinant of model performance.

In single-agent settings, a rich prompt optimization toolkit already exists. ADePT~\citep{tang2025adeptadaptivedecomposedprompt} proposes adaptive decomposed prompt tuning to improve stability. GRACE~\citep{shi2025lossgaingatedrefinement} introduces gated refinement with rejection control to avoid regression. DynaPrompt~\citep{xiao2025dynapromptdynamictesttimeprompt} handles online prompt drift with dynamic buffers. In multi-agent settings, the problem becomes more complex because agent interactions and collaboration structure must also be considered. HiveMind~\citep{xia2025hivemindcontributionguidedonlineprompt} attributes each agent's contribution through approximate Shapley values and rewrites underperforming prompts online. MAPRO~\citep{zhang2025maprorecastingmultiagentprompt} formulates multi-agent prompt optimization as MAP inference over a directed acyclic graph. MASS~\citep{zhou2026multiagentdesignoptimizingagents} shows that prompt optimization and topology optimization are interdependent through alternating optimization. Our work differs from these methods in problem formulation. Prior work optimizes prompts either for individual agents or within a fixed multi-agent structure. In contrast, our framework uses routing-induced diagnostic signals to identify underperforming roles, refine their prompts, and retrain the router on the updated agent pool.

\subsection{Knowledge Graph Question Answering and Graph Neural Networks}
Knowledge Graph Question Answering (KGQA) has undergone significant advancements, evolving from early approaches such as semantic parsing and retrieval-based methods. Initial models translated natural language queries into structured formats while more recent progress integrates large language models (LLMs) to improve both retrieval efficiency and reasoning ability. Approaches like Jiang et al. \citep{jiang2023structgpt} and Wang et al. \citep{wang2023knowledgpt} utilize LLMs to transform queries into formats such as SQL or SPARQL~\citep{li2025graph}, enhancing retrieval accuracy. Others, such as Kim et al. \citep{kim2023kg} and Gao et al. \citep{gao2024two}, focus on reasoning over retrieved subgraphs or triples, tackling multi-hop reasoning tasks in KGQA. However, most benchmarks in this field are designed for general-purpose datasets and fail to address domain-specific complexities, such as the challenges unique to nutritional health reasoning and multi-hop medical reasoning~\citep{huang2024ritek}. To learn the KG representation for downstream QA, a common practice is to use graph neural networks (GNN).

GNNs are a class of learning models specifically designed to operate on graph-structured data and have demonstrated substantial success across a variety of domains \citep{kipf2016semi, velivckovic2017graph, hamilton2017inductive, ma2025llm}. They have been effectively applied in social network analysis, recommendation systems, biological interaction networks, and molecular property prediction, among others, by leveraging the relational inductive biases inherent in graph structures. A key strength of GNNs lies in their ability to generalize across graph instances of varying sizes and topologies, enabling their deployment in diverse and dynamic real-world settings~\citep{ma2023hypergraph, ma2025adaptive,DBLP:journals/corr/abs-2601-22384}. Moreover, recent efforts have explored the transferability of GNNs across tasks and domains \citep{wang2024subgraph, cao2023pre, zhang2024diet}, including pretraining strategies and task-agnostic embeddings, which draw inspiration from the success of transfer learning in language and vision domains. However, challenges such as oversmoothing, limited expressiveness, and poor scalability remain active research areas. Looking ahead, the field is shifting toward the development of graph foundation models: large-scale, pretrained GNNs designed to capture generalizable structural and semantic patterns across graph corpora \citep{wang2024gft, wang2025neural, wang2024learning}. These models aim to provide reusable, adaptable representations and serve as backbones for a wide range of downstream tasks, mirroring the transformative impact of foundation models in NLP and vision.

\section{Additional Method Details}

\subsection{Details of Prompt Refinement}

\noindent\textbf{Role-level consistency filter.}
Before ranking agents for prompt refinement, we apply a role-level consistency filter to distinguish prompt-level weaknesses from backbone-specific outliers. For each role $r$, we examine its relative performance across backbones and retain the role only if it underperforms on at least $\tau_{\mathrm{bb}}$ backbones. In our implementation, we first set $\tau_{\mathrm{bb}}=3$; if this yields no candidate roles, we relax it to $2$, and then to $1$ if necessary. This filter ensures that refinement is directed toward roles with consistent weaknesses rather than isolated backbone-specific failures.

\noindent\textbf{Priority scoring.}
For each surviving backbone--role pair $a$, we compute
\[
\mathrm{priority}(a)=\mathrm{severity}(a)\cdot(\alpha+\bar{w}(a)),
\]
where $\mathrm{severity}(a)=1-\bar{\mathrm{F1}}(a)$, $\bar{w}(a)$ is the router's average learned weight assigned to $a$, and $\alpha=0.3$ is a fixed base coefficient. In our implementation, $\bar{w}(a)$ is computed by averaging routing weights aggregated by question category. This score prioritizes agents that both underperform and remain important to the router.

  \begin{algorithm*}[t]
  \caption{\textsc{EvolveRouter}: Co-Optimizing Routing and Prompts}
  \label{alg:overall}
  \begin{algorithmic}[1]
  \Require Agent pool $\mathcal{A}$ with prompts $\{\pi_a\}$; train/val splits
  $\mathcal{D}_{\text{train}}, \mathcal{D}_{\text{val}}$; refinement rounds $R$
  \Ensure Router parameters $\theta^*$; refined prompts $\{\pi_a^*\}$

  \Statex \hspace{-1.2em} \textbf{Stage 1: Router Warm-up and Diagnostics}
  \State Execute all agents on $\mathcal{D}_{\text{train}}$; compute F1 scores
  $\{\text{F1}_a(q)\}$
  \State Convert to soft targets: $p^*(a \mid q) \propto \exp(\text{F1}_a(q) /
  \tau_{\text{temp}})$
  \State Train RouterGNN $\theta^{(0)}$ by minimizing
  $\mathcal{L}_{\text{KL}}(p^* \| p_\theta)$
  \State Collect diagnostics: failure archive, per-agent severity, router weight
   statistics

  \Statex
  \Statex \hspace{-1.2em} \textbf{Stage 2: Closed-Loop Prompt Refinement}
  \For{$r = 1, \dots, R$}
      \State \textit{Target selection:} rank agents by $\text{priority}(a) = (1
  - \overline{\text{F1}}(a)) \cdot (\alpha + \bar{w}(a))$, filtered for
  cross-backbone consistency
      \For{each top-$N_{\max}$ target agent $a$}
          \State Generate $N_c$ candidate prompts via constrained rewriting from
   failure evidence
          \State Evaluate candidates on targeted validation subset
  $D_a^{\text{val}}$
          \State Accept best candidate if it passes the validation gate;
  otherwise freeze after repeated failures
      \EndFor
      \State Re-query modified agents; recompute soft targets with updated
  outputs
      \State Fine-tune $\theta^{(r)}$ from $\theta^{(r-1)}$; revert if
  validation F1 regresses
  \EndFor

  \Statex
  \Statex \hspace{-1.2em} \textbf{Stage 3: Adaptive Inference}
  \For{each test query $q$}
      \State Rank agents by $p_{\theta^*}(a \mid q, \mathcal{G})$ in descending
  order $\sigma$
      \For{$k = 1, \dots, k_{\max}$}
          \State Query agent $a_{\sigma(k)}$; compute weighted agreement $A(k,
  q)$
          \State \textbf{if} $k \geq k_{\min}$ and $A(k, q) \geq \tau$
  \textbf{then break}
      \EndFor
      \State \textbf{return} weighted majority vote over consulted agents
  \EndFor
  \end{algorithmic}
  \end{algorithm*}
  
\noindent\textbf{Target budget and freezing.}
In each round, we optimize only the top-$N_{\max}$ agents by priority, with $N_{\max}=3$ in our experiments. Agents that experience repeated failed rewrite attempts are frozen and removed from future rounds. Specifically, after $k_{\mathrm{freeze}}=3$ consecutive rejections, the agent is marked as saturated.

\noindent\textbf{Mutation strategy and candidate generation.}
For each selected agent, we generate $N_c=3$ candidate rewrites at temperatures $\{0.30, 0.45, 0.60\}$. The rewriter receives the current prompt, representative failure cases, and error-pattern statistics. To prevent prompt drift, the rewriter is instructed to preserve the role's core reasoning strategy and output format, and may only add, rephrase, or remove a small number of sentences. In practice, this corresponds to adding 1--3 clarifying sentences, rephrasing for clarity, or removing misleading instructions.

\noindent\textbf{Targeted validation subset.}
Each candidate prompt is evaluated on a targeted validation subset centered on the diagnosed failures of the target agent and supplemented with additional validation samples. This subset design keeps validation focused on the identified weaknesses while reducing overfitting to a narrow slice of errors. Among the candidates, we retain the one with the best validation performance. This is a more robust variant of the earlier failure-only validation design used in preliminary drafts.

\noindent\textbf{Rejection gate.}
The acceptance rule used in code is more permissive than a strict ``new mean must be larger than old mean'' criterion. Let $\bar{\mathrm{F1}}(\pi_a)$ denote the mean F1 of the current prompt on the targeted validation subset, and let $n_{\uparrow}$ and $n_{\downarrow}$ denote the numbers of validation samples whose F1 improves or degrades under the candidate prompt. A candidate rewrite is accepted if it does not regress by more than $\delta$ on mean F1, changes at least one sample, and does not introduce net degradation:
\[
\pi_a \leftarrow \pi_a^\star
\quad \text{if} \quad
\bar{\mathrm{F1}}(\pi_a^\star) \ge \bar{\mathrm{F1}}(\pi_a)-\delta
\;\wedge\;
n_{\uparrow} > 0
\;\wedge\;
\neg\Bigl(
\bar{\mathrm{F1}}(\pi_a^\star) < \bar{\mathrm{F1}}(\pi_a)
\;\wedge\;
n_{\downarrow} > n_{\uparrow}
\Bigr).
\]
This rule allows small mean drops within tolerance when offset by broader sample-level improvements, while still blocking clearly harmful rewrites.

\noindent\textbf{Incremental router update.}
After accepted prompt revisions, the router is updated by resuming from the previous checkpoint rather than retraining from random initialization. Round 1 resumes from the Stage 1 warm-up checkpoint, and later rounds resume from the previous round. We found that retraining from scratch consistently discards useful routing knowledge and degrades performance. The update is made efficient through cache seeding: the answer cache from the previous round is copied forward, and because cache keys include a prompt hash, unchanged agents reuse cached outputs while modified agents are automatically invalidated and re-queried. In practice, only 2--3 out of 24 agents are modified per round, so the LLM cost is roughly proportional to the number of changed agents rather than the full pool. 

\noindent\textbf{Regression check and multi-round chaining.}
After router fine-tuning, we apply a regression check on validation F1 and revert the update if the new router falls more than $0.03$ below the previous checkpoint. Prompt refinement is chained across rounds. Each round uses the previous round's checkpoint, diagnostics, and seeded cache, so optimization targets evolve as earlier weaknesses are corrected. In practice, improvement usually saturates after 2--3 rounds, and the process terminates early when no candidate prompt passes validation or when all remaining candidates are frozen.

\section{Implementation Details}
\label{app:impl}
\subsection{Benchmarks}
\label{app:benchmarks}

\textbf{HotpotQA}~\citep{yang2018hotpotqadatasetdiverseexplainable} targets multi-hop reasoning over Wikipedia, containing more than 100,000 questions that each require aggregating evidence from at least two separate paragraphs. Every instance is accompanied by annotated supporting facts, enabling fine-grained evaluation of whether a model identifies the correct reasoning chain rather than arriving at the answer by shortcut. The need to traverse multiple evidence sources makes HotpotQA a natural fit for testing whether routers can learn which agent strategies excel at cross-document inference.

\textbf{2WikiMultihopQA}~\citep{ho2020constructingmultihopqadataset} raises the difficulty of multi-hop QA by systematically pairing entities from two distinct Wikipedia articles and generating compositional questions whose answers cannot be deduced from either article alone. The dataset covers several structured reasoning patterns---comparison, bridge, and inference---and its controlled construction guarantees that solving a question genuinely requires evidence fusion. This makes it a particularly strict benchmark for evaluating how well routing mechanisms handle diverse compositional structures.

\textbf{NewsQA}~\citep{trischler2017newsqamachinecomprehensiondataset} draws from over 10,000 CNN articles and contains roughly 120,000 question--answer pairs. Crowdworkers wrote questions after reading only the headline, so questions are genuinely information-seeking and often require locating evidence scattered across long narrative paragraphs. The dataset also includes unanswerable and ambiguous items, introducing a realistic challenge: an effective router must distinguish agents that are robust to noise and longer contexts from those that rely on surface-level pattern matching.

\textbf{TriviaQA}~\citep{joshi2017triviaqalargescaledistantly} collects over 95,000 trivia questions from enthusiast websites and pairs them with evidence documents harvested independently from the web and Wikipedia. Because questions and documents are authored separately, lexical overlap between the query and the answer passage is low, demanding genuine semantic comprehension rather than keyword matching. The noisy, distantly supervised nature of the evidence further tests whether routing can identify agents that remain accurate under imperfect retrieval conditions.

\textbf{NGQA}~\citep{zhang2024ngqanutritionalgraphquestion} shifts the evaluation to a specialized health domain. Built on NHANES dietary survey data and the FNDDS nutrient database, NGQA asks whether a given food is appropriate for a particular user's health profile by reasoning over nutritional attributes, medical conditions, and dietary constraints. Three complexity tiers---\emph{sparse}, \emph{standard}, and \emph{complex}---vary the amount of contextual information available per question. This benchmark probes whether routing frameworks trained on general QA can transfer to domains with structured, domain-specific entity interactions.

\begin{table*}[t]
\centering
\renewcommand{\arraystretch}{1.1}
\resizebox{0.9\textwidth}{!}{
\begin{tabular}{llcccccc}
\toprule
\textbf{Benchmark} & \textbf{Agent} & \textbf{Raw} & \textbf{CoT} & \textbf{SC} & \textbf{React Reflect} & \textbf{MAD} & \textbf{Summary} \\
\midrule

\multirow{4}{*}{\textbf{NewsQA}} 
& Qwen2.5-7B     & 60.51{\tiny$\pm$0.52} & 59.51{\tiny$\pm$0.96} & 58.84{\tiny$\pm$0.21} & 59.74{\tiny$\pm$0.63} & 59.80{\tiny$\pm$0.42} & 60.43{\tiny$\pm$0.48} \\
& gpt-oss-20B    & 60.55{\tiny$\pm$1.45} & 61.29{\tiny$\pm$0.66} & 58.91{\tiny$\pm$1.35} & 59.89{\tiny$\pm$1.02} & 62.80{\tiny$\pm$2.15} & 61.43{\tiny$\pm$0.78} \\
& Mixtral-8x7B   & 53.79{\tiny$\pm$0.99} & 54.03{\tiny$\pm$0.84} & 55.69{\tiny$\pm$1.23} & 54.46{\tiny$\pm$0.66} & 55.13{\tiny$\pm$0.51} & 55.16{\tiny$\pm$1.05} \\
& Llama-3-8B     & 55.50{\tiny$\pm$0.78} & 56.24{\tiny$\pm$0.54} & 57.97{\tiny$\pm$0.87} & 56.55{\tiny$\pm$0.81} & 57.07{\tiny$\pm$0.93} & 55.97{\tiny$\pm$0.72} \\
\midrule

\multirow{4}{*}{\textbf{HotpotQA}} 
& Qwen2.5-7B     & 59.38{\tiny$\pm$1.21} & 58.04{\tiny$\pm$1.81} & 58.09{\tiny$\pm$1.04} & 56.89{\tiny$\pm$1.42} & 56.12{\tiny$\pm$1.37} & 57.24{\tiny$\pm$1.53} \\
& gpt-oss-20B    & 68.19{\tiny$\pm$1.37} & 68.74{\tiny$\pm$1.39} & 67.04{\tiny$\pm$1.02} & 68.39{\tiny$\pm$1.26} & 67.74{\tiny$\pm$0.98} & 66.63{\tiny$\pm$1.14} \\
& Mixtral-8x7B   & 58.40{\tiny$\pm$0.79} & 57.46{\tiny$\pm$0.96} & 53.18{\tiny$\pm$1.04} & 58.22{\tiny$\pm$1.21} & 58.88{\tiny$\pm$0.71} & 58.98{\tiny$\pm$0.87} \\
& Llama-3-8B     & 56.19{\tiny$\pm$0.93} & 55.26{\tiny$\pm$0.69} & 55.83{\tiny$\pm$0.71} & 56.00{\tiny$\pm$0.77} & 54.97{\tiny$\pm$0.85} & 52.63{\tiny$\pm$0.56} \\
\midrule

\multirow{4}{*}{\textbf{2Wiki}} 
& Qwen2.5-7B     & 49.52{\tiny$\pm$2.58} & 47.12{\tiny$\pm$1.21} & 46.27{\tiny$\pm$1.35} & 51.02{\tiny$\pm$2.37} & 47.44{\tiny$\pm$1.27} & 50.67{\tiny$\pm$2.11} \\
& gpt-oss-20B    & 71.77{\tiny$\pm$1.71} & 72.70{\tiny$\pm$2.07} & 71.70{\tiny$\pm$0.42} & 74.47{\tiny$\pm$0.67} & 75.21{\tiny$\pm$1.11} & 72.65{\tiny$\pm$1.24} \\
& Mixtral-8x7B   & 57.95{\tiny$\pm$2.07} & 55.88{\tiny$\pm$1.76} & 51.13{\tiny$\pm$1.39} & 51.94{\tiny$\pm$1.79} & 55.84{\tiny$\pm$1.54} & 52.61{\tiny$\pm$0.89} \\
& Llama-3-8B     & 46.46{\tiny$\pm$1.77} & 47.07{\tiny$\pm$1.66} & 48.46{\tiny$\pm$1.61} & 45.40{\tiny$\pm$1.40} & 44.48{\tiny$\pm$0.18} & 48.82{\tiny$\pm$1.78} \\
\midrule

\multirow{4}{*}{\textbf{TriviaQA}} 
& Qwen2.5-7B     & 36.44{\tiny$\pm$1.00} & 39.22{\tiny$\pm$2.51} & 37.03{\tiny$\pm$1.97} & 41.43{\tiny$\pm$2.35} & 38.88{\tiny$\pm$1.09} & 36.71{\tiny$\pm$2.72} \\
& gpt-oss-20B    & 37.38{\tiny$\pm$2.11} & 45.28{\tiny$\pm$2.43} & 42.76{\tiny$\pm$2.10} & 43.03{\tiny$\pm$1.19} & 42.66{\tiny$\pm$1.43} & 43.83{\tiny$\pm$2.77} \\
& Mixtral-8x7B   & 51.41{\tiny$\pm$0.65} & 60.27{\tiny$\pm$2.52} & 60.10{\tiny$\pm$1.77} & 59.56{\tiny$\pm$1.92} & 60.19{\tiny$\pm$1.98} & 59.49{\tiny$\pm$2.40} \\
& Llama-3-8B     & 41.94{\tiny$\pm$2.17} & 46.60{\tiny$\pm$1.97} & 51.77{\tiny$\pm$1.70} & 47.50{\tiny$\pm$1.22} & 49.02{\tiny$\pm$2.36} & 50.17{\tiny$\pm$1.27} \\
\midrule

\multirow{4}{*}{\textbf{NGQA}} 
& Qwen2.5-7B     & 46.31{\tiny$\pm$0.60} & 46.82{\tiny$\pm$0.42} & 45.71{\tiny$\pm$0.55} & 46.21{\tiny$\pm$2.36} & 43.90{\tiny$\pm$0.52} & 48.31{\tiny$\pm$0.43} \\
& gpt-oss-20B    & 33.70{\tiny$\pm$0.94} & 35.70{\tiny$\pm$0.28} & 34.53{\tiny$\pm$0.99} & 34.22{\tiny$\pm$0.94} & 34.71{\tiny$\pm$1.04} & 34.50{\tiny$\pm$1.79} \\
& Mixtral-8x7B   & 42.02{\tiny$\pm$1.65} & 40.91{\tiny$\pm$0.28} & 41.90{\tiny$\pm$0.46} & 41.90{\tiny$\pm$1.13} & 42.62{\tiny$\pm$2.45} & 41.91{\tiny$\pm$1.93} \\
& Llama-3-8B     & 48.03{\tiny$\pm$0.74} & 45.91{\tiny$\pm$0.69} & 48.12{\tiny$\pm$0.66} & 48.41{\tiny$\pm$0.71} & 47.20{\tiny$\pm$0.39} & 45.62{\tiny$\pm$0.53} \\
\bottomrule
\end{tabular}}
\caption{F1 scores across five benchmarks for different agents under various prompting strategies. We report mean and standard deviation.}
\label{tab:raw_f1}
\vspace{-20pt}
\end{table*}
\subsection{Baselines}
\label{app:baselines}

\textbf{Agent designs.}\quad We assemble six prompting strategies that span a spectrum from single-turn generation to coordinated multi-agent reasoning. \emph{Raw} queries the backbone directly without any scaffolding. \emph{Chain-of-Thought}~\citep{wei2022chain} instructs the model to produce step-by-step reasoning before answering. \emph{Self-Consistency}~\citep{wang2023selfconsistencyimproveschainthought} samples several CoT trajectories and returns the answer with the highest agreement, trading compute for reduced variance. \emph{ReAct-Reflection}~\citep{yao2022react,shinn2023reflexion} interleaves reasoning with self-evaluation, allowing the agent to revise its answer when it detects inconsistencies. \emph{Multi-Agent Debate}~\citep{du2024improving} simulates an adversarial discussion among debaters and a judge to converge on a consensus answer. \emph{Multi-Agent Summary} asks parallel reasoners to contribute partial analyses, which are then distilled into a single response. Each strategy interacts differently with the context, so performance rankings vary across tasks and backbones---precisely the heterogeneity that motivates learned routing.

\textbf{Heuristic ensembling.}\quad Four non-learned baselines establish reference points. \emph{Average} reports the mean metric across all 24 agents, approximating uniform random selection. \emph{Majority Vote} canonicalizes every agent answer and picks the most frequent one, leveraging consensus without any weighting. \emph{Best LLM} averages scores within each backbone and reports the top one, reflecting the ceiling of backbone-only selection. \emph{Best Agent} selects the single strongest backbone--role pair per dataset, representing the best achievable performance without any collaboration.

\textbf{Training-free routing.}\quad \emph{Random Router} draws an agent uniformly at random for each query (10 seeds, mean$\pm$std reported). \emph{CascadeRouting}~\citep{dekoninck2025unifiedapproachroutingcascading} trains a lightweight quality estimator with binary cross-entropy labels and calibrates a temperature on the validation set via grid search. \emph{RouteLLM}~\citep{ong2025routellmlearningroutellms} projects sentence embeddings (all-MiniLM-L6-v2, 384-d) into a 64-d latent space and scores agents via dot product with learnable agent vectors, trained with MSE loss.

\textbf{Adaptive routing.}\quad \emph{LLM-Blender}~\citep{jiang-etal-2023-llm} deduplicates all 24 agent outputs and asks a meta-LLM (Llama-3.3-70B-Instruct) to select the best candidate. \emph{HybridLLM}~\citep{ding2024hybridllmcostefficientqualityaware} first checks whether agents reach sufficient consensus; if so, the majority answer is accepted without an LLM call, otherwise the meta-LLM adjudicates. \emph{KNN Router}~\citep{hu2024routerbenchbenchmarkmultillmrouting} finds the 5 nearest training questions by cosine similarity and selects the agent with the highest mean F1 among neighbors. \emph{MLP Router}~\citep{hu2024routerbenchbenchmarkmultillmrouting} is a feedforward classifier trained with cross-entropy to map question embeddings to agent labels.

\textbf{Structure-aware routing.}\quad \emph{GraphRouter}~\citep{feng2025graphroutergraphbasedrouterllm} treats routing as heterogeneous link prediction, encoding query--model interactions within a graph and using GNNs to learn compatibility scores from the resulting topology. By explicitly capturing relational patterns between inputs and agents, it establishes a competitive structure-aware baseline for learned routing. \emph{AgentRouter}~\citep{zhang2025agentrouterknowledgegraphguidedllmrouter} further integrates contextual entities into the graph and trains a heterogeneous GNN with soft F1-based supervision via KL divergence. An \emph{Oracle} that always picks the per-query best agent provides the theoretical ceiling.

\begin{table*}[t]
\centering
\renewcommand{\arraystretch}{1.15}
\resizebox{0.9\textwidth}{!}{
\begin{tabular}{llcccccc}
\toprule
\textbf{Benchmark} & \textbf{Agent} & \textbf{Raw} & \textbf{CoT} & \textbf{SC} & \textbf{React Reflect} & \textbf{MAD} & \textbf{Summary} \\
\midrule

\multirow{4}{*}{\textbf{NewsQA}} 
& Qwen2.5-7B     & 41.33{\tiny$\pm$0.58} & 39.00{\tiny$\pm$0.00} & 42.00{\tiny$\pm$0.00} & 42.33{\tiny$\pm$0.58} & 41.33{\tiny$\pm$0.58} & 42.00{\tiny$\pm$0.00} \\
& gpt-oss-20B    & 39.80{\tiny$\pm$2.05} & 37.33{\tiny$\pm$1.73} & 35.00{\tiny$\pm$2.00} & 35.67{\tiny$\pm$1.53} & 37.11{\tiny$\pm$1.15} & 37.67{\tiny$\pm$1.53} \\
& Mixtral-8x7B   & 28.33{\tiny$\pm$0.58} & 28.00{\tiny$\pm$1.00} & 28.67{\tiny$\pm$0.58} & 29.00{\tiny$\pm$0.00} & 28.00{\tiny$\pm$1.00} & 27.33{\tiny$\pm$0.58} \\
& Llama-3-8B     & 37.33{\tiny$\pm$1.15} & 38.00{\tiny$\pm$1.00} & 39.67{\tiny$\pm$0.58} & 37.33{\tiny$\pm$1.15} & 39.00{\tiny$\pm$0.00} & 38.00{\tiny$\pm$1.00} \\
\midrule

\multirow{4}{*}{\textbf{HotpotQA}} 
& Qwen2.5-7B     & 47.33{\tiny$\pm$0.58} & 46.33{\tiny$\pm$0.58} & 46.00{\tiny$\pm$1.00} & 44.00{\tiny$\pm$0.00} & 42.67{\tiny$\pm$1.15} & 44.00{\tiny$\pm$1.00} \\
& gpt-oss-20B    & 51.79{\tiny$\pm$2.09} & 56.15{\tiny$\pm$0.87} & 54.00{\tiny$\pm$1.00} & 54.67{\tiny$\pm$1.15} & 55.67{\tiny$\pm$1.15} & 53.00{\tiny$\pm$1.00} \\
& Mixtral-8x7B   & 45.33{\tiny$\pm$0.58} & 44.33{\tiny$\pm$1.15} & 40.00{\tiny$\pm$1.00} & 45.00{\tiny$\pm$1.00} & 45.67{\tiny$\pm$1.15} & 44.00{\tiny$\pm$0.00} \\
& Llama-3-8B     & 44.33{\tiny$\pm$0.58} & 42.00{\tiny$\pm$1.00} & 44.67{\tiny$\pm$0.58} & 43.00{\tiny$\pm$1.00} & 41.67{\tiny$\pm$1.15} & 41.33{\tiny$\pm$1.15} \\
\midrule

\multirow{4}{*}{\textbf{2Wiki}} 
& Qwen2.5-7B     & 41.33{\tiny$\pm$0.58} & 38.00{\tiny$\pm$1.00} & 36.00{\tiny$\pm$1.00} & 42.33{\tiny$\pm$0.58} & 39.33{\tiny$\pm$0.58} & 42.00{\tiny$\pm$1.73} \\
& gpt-oss-20B    & 62.72{\tiny$\pm$1.32} & 63.67{\tiny$\pm$1.15} & 62.00{\tiny$\pm$1.73} & 64.33{\tiny$\pm$1.15} & 64.56{\tiny$\pm$1.91} & 63.67{\tiny$\pm$1.15} \\
& Mixtral-8x7B   & 46.33{\tiny$\pm$0.58} & 44.33{\tiny$\pm$0.58} & 40.67{\tiny$\pm$1.15} & 40.00{\tiny$\pm$1.00} & 43.00{\tiny$\pm$1.00} & 40.33{\tiny$\pm$1.15} \\
& Llama-3-8B     & 39.00{\tiny$\pm$1.00} & 39.33{\tiny$\pm$0.58} & 41.33{\tiny$\pm$0.58} & 36.00{\tiny$\pm$0.00} & 36.67{\tiny$\pm$0.58} & 41.33{\tiny$\pm$1.15} \\
\midrule

\multirow{4}{*}{\textbf{TriviaQA}} 
& Qwen2.5-7B     & 32.33{\tiny$\pm$0.58} & 34.33{\tiny$\pm$0.58} & 33.00{\tiny$\pm$0.00} & 34.67{\tiny$\pm$0.58} & 32.33{\tiny$\pm$0.58} & 30.00{\tiny$\pm$1.00} \\
& gpt-oss-20B    & 30.00{\tiny$\pm$1.00} & 37.00{\tiny$\pm$1.00} & 33.67{\tiny$\pm$1.15} & 34.67{\tiny$\pm$0.58} & 33.33{\tiny$\pm$1.15} & 34.00{\tiny$\pm$1.00} \\
& Mixtral-8x7B   & 42.09{\tiny$\pm$0.59} & 50.13{\tiny$\pm$2.01} & 51.00{\tiny$\pm$2.00} & 49.33{\tiny$\pm$1.53} & 49.00{\tiny$\pm$1.00} & 50.67{\tiny$\pm$1.15} \\
& Llama-3-8B     & 35.33{\tiny$\pm$0.58} & 40.33{\tiny$\pm$0.58} & 43.67{\tiny$\pm$1.15} & 40.00{\tiny$\pm$1.00} & 42.33{\tiny$\pm$1.15} & 42.00{\tiny$\pm$0.00} \\

\midrule

\multirow{4}{*}{\textbf{NGQA}} 
& Qwen2.5-7B     & 5.00{\tiny$\pm$1.00} & 4.33{\tiny$\pm$0.58} & 2.00{\tiny$\pm$1.00} & 2.33{\tiny$\pm$0.58} & 2.33{\tiny$\pm$0.58} & 4.00{\tiny$\pm$1.00} \\
& gpt-oss-20B    & 2.33{\tiny$\pm$0.58} & 3.00{\tiny$\pm$1.00} & 4.00{\tiny$\pm$1.00} & 3.67{\tiny$\pm$1.15} & 4.33{\tiny$\pm$0.58} & 3.33{\tiny$\pm$0.58} \\
& Mixtral-8x7B   & 5.33{\tiny$\pm$0.58} & 5.00{\tiny$\pm$0.00} & 5.00{\tiny$\pm$0.00} & 5.33{\tiny$\pm$0.58} & 4.00{\tiny$\pm$1.00} & 5.33{\tiny$\pm$1.53} \\
& Llama-3-8B     & 5.00{\tiny$\pm$2.00} & 4.00{\tiny$\pm$1.00} & 7.67{\tiny$\pm$1.53} & 5.67{\tiny$\pm$1.53} & 4.00{\tiny$\pm$1.00} & 2.33{\tiny$\pm$1.15} \\
\bottomrule
\end{tabular}}
\caption{Exact Match (EM) scores across benchmarks for different agents under various prompting strategies.}
\label{tab:raw_em}
\vspace{-20pt}
\end{table*}

\subsection{Training Details}
\label{app:training}
\textbf{Data splits.}\quad From each benchmark we sample 500 training, 100 validation, and 100 test instances. Every instance is transformed into a heterogeneous knowledge graph: entity nodes are extracted with spaCy NER, entity--entity edges are derived from dependency-parsed relation triples with intervening relation nodes, query--entity edges capture explicit mentions, and agent--entity edges record which entities each backbone deems most relevant. Query--agent edges are left as trainable parameters. Node features are initialized in a shared textual embedding space. We evaluate with token-level F1 and Exact Match (EM) following SQuAD conventions~\citep{rajpurkar2016squad100000questionsmachine}, and average all results over three seeds.

\textbf{Router architecture.}\quad RouterGNN stacks $L{=}2$ HGTConv layers with hidden dimension 256 and dropout 0.2, applying instance normalization per node type. After message passing, routing scores are produced by a two-layer MLP operating on the concatenated query and agent representations. An auxiliary head predicts the question type with label smoothing 0.05.

\textbf{Optimization.}\quad Adam with learning rate $10^{-4}$; soft-target temperature $\tau{=}0.25$; label smoothing $\epsilon{=}10^{-3}$; backbone-entropy regularization $\lambda_\text{bb}{=}0.02$; agent-entropy regularization $\lambda_\text{ent}{=}10^{-3}$. Training runs up to 20 epochs and stops early after 5 epochs without validation F1 improvement.

\textbf{Prompt optimization.}\quad We use GPT-4o-mini as the rewriter LLM. Each co-evolution round retrains the router for 15 epochs from the preceding checkpoint. Per selected agent, three candidate rewrites are sampled at temperatures $\{0.30, 0.45, 0.60\}$. Up to $K$ agents are refined per round; agents with 3 consecutive failed rewrites are frozen. A candidate is accepted only if it does not regress validation F1. Improvement generally plateaus after 2--3 rounds.

\textbf{Agent pool and caching.}\quad Four backbones (Llama-3-8B-Instruct, Qwen2.5-7B-Instruct-Turbo, Mixtral-8$\times$7B-v0.1, and gpt-oss-20b) are paired with six roles to form 24 agents. LLM calls use temperature 0.2 and a 3{,}000-token output cap. Answers are cached with composite keys that include a hash of the active prompt, so modifying a prompt during refinement automatically invalidates the affected entries while preserving outputs for unchanged agents.

\textbf{Computational cost.}\quad Training (Stage~1) requires $500 \times 24 = 12{,}000$ LLM calls to collect per-agent answers. Each refinement round modifies only 2--3 agents, keeping the incremental cost at roughly 10\% of the full evaluation. The GNN training itself takes under 5 minutes per epoch on a single GPU. At inference time, Adaptive~K reduces the number of agent calls from $n{=}24$ to an average of 2.9--7.4 across our benchmarks, yielding a 69--88\% reduction while maintaining or improving F1.

\section{Additional Experiments}
\label{app:additional}
\subsection{Application to Nutritional Health Domain: A Cost-Effective Perspective}
\label{app:glen}
\begin{wraptable}{r}{0.55\textwidth}
\vspace{-12pt}
\centering
\small
\caption{Ablation on GLEN-Bench and NGQA. Both benchmarks target nutritional QA but differ in graph density and contextual richness.}
\resizebox{\linewidth}{!}{
\begin{tabular}{l cc cc}
\toprule
\multirow{2}{*}{\textbf{Method}} &
\multicolumn{2}{c}{\textbf{GLEN-Bench}} &
\multicolumn{2}{c}{\textbf{NGQA}} \\
\cmidrule(lr){2-3}\cmidrule(lr){4-5}
& \textbf{F1} & \textbf{EM}
& \textbf{F1} & \textbf{EM} \\
\toprule
KG Router (base)
&49.39{\tiny$\pm$0.81} &11.00{\tiny$\pm$1.00} 
&49.95{\tiny$\pm$1.29}&8.67{\tiny$\pm$1.53}
\\
\rowcolor{gray!20}
\multicolumn{5}{c}{\textit{Prompt Optimization Ablation}} \\
+PO (Round 1) 
&50.97{\tiny$\pm$0.46} & 11.33{\tiny$\pm$0.58}
&50.31{\tiny$\pm$0.88}&9.00{\tiny$\pm$0.00}
\\
+PO (Round 2)
&51.74{\tiny$\pm$0.38} & 12.00{\tiny$\pm$1.00}
&52.99{\tiny$\pm$1.19}&9.33{\tiny$\pm$0.58}
\\
+PO (Round 3)
&52.28{\tiny$\pm$0.86} & 14.33{\tiny$\pm$0.58}
&53.28{\tiny$\pm$0.16}&9.67{\tiny$\pm$0.58}
\\
\rowcolor[RGB]{222,230,241}
\textbf{$\Delta$ vs.\ base (\%)} 
&$\uparrow$5.85\% &$\uparrow$30.27\% & $\uparrow$ 6.67\% &$\uparrow$ 11.53\%\\
\midrule
\rowcolor{gray!20}
\multicolumn{5}{c}{\textit{Adaptive K Ablation}} \\
+ Adaptive K
&53.46{\tiny$\pm$1.41} &14.00{\tiny$\pm$1.00} 
&50.30{\tiny$\pm$1.74}&9.00{\tiny$\pm$1.00}
\\
\rowcolor[RGB]{222,230,241}
\textbf{$\Delta$ vs.\ base (\%)} 
&$\uparrow$8.24\% &$\uparrow$27.27\% &$\uparrow$ 0.70\%&$\uparrow$ 3.81\%
\\
\midrule
Fixed K=3
&51.41{\tiny$\pm$1.73}&13.33{\tiny$\pm$1.15}
&49.76{\tiny$\pm$0.49}&8.67{\tiny$\pm$0.58}
\\
Fixed K=5
&52.09{\tiny$\pm$1.15}&13.67{\tiny$\pm$1.15}
&48.94{\tiny$\pm$0.59}&7.33{\tiny$\pm$0.58}
\\
Fixed K=10
&52.42{\tiny$\pm$0.91}&14.00{\tiny$\pm$1.00}
&48.34{\tiny$\pm$0.31}&7.00{\tiny$\pm$0.00}
\\
\midrule
\rowcolor[RGB]{222,230,241}
\textbf{EvolveRouter} 
&55.16{\tiny$\pm$0.39} &16.33{\tiny$\pm$0.58} &53.28{\tiny$\pm$0.16}& 9.67{\tiny$\pm$0.58}
\\
\bottomrule
\end{tabular}
}
\label{tab:glen}
\vspace{-8pt}
\end{wraptable}
With the growing awareness of the importance of dietary health, various studies have sought to incorporate health metrics into applications such as health-aware food recommendation or nutritional health question answering \citep{li2024cheffusion, zhang2025mopi}. To further evaluate our framework on this domain beyond NGQA, we conduct experiments on GLEN-Bench~\citep{huang2026glenbenchgraphlanguagebasedbenchmark}, a recently introduced multi-task benchmark for nutritional health assessment that not only includes health information but also the affordability of the foods. GLEN-Bench is constructed by integrating three large-scale population-level data sources: NHANES (2003--2020) demographic and clinical records, the FNDDS nutrient database, and the USDA Purchase-to-Plate crosswalk. While NGQA focuses on nutritional reasoning through a relatively sparse graph built from food--nutrient--condition triplets, GLEN-Bench assembles a substantially denser heterogeneous knowledge graph that additionally incorporates \textbf{food price information} (discretized into affordability tiers from USDA pricing data), \textbf{socioeconomic indicators} such as poverty status and food insecurity, and \textbf{dietary habit profiles} extracted from 48 NHANES questionnaire categories. This richer graph structure (across 9 node types and 11 relation types) poses a distinct challenge for the router: the contextual information per query is considerably denser and more heterogeneous than in NGQA, requiring the model to navigate a broader space of entity interactions.

Table~\ref{tab:glen} presents the results. Despite the substantially denser graph structure, the overall trends are consistent with those observed on the other benchmarks. PO yields cumulative gains on both datasets ($\uparrow$5.85\% F1 on GLEN-Bench, $\uparrow$6.67\% on NGQA after three rounds), and the full EvolveRouter achieves the best result in each case. One notable difference is Adaptive K, which contributes a much larger improvement on GLEN-Bench ($\uparrow$8.24\%) than on NGQA ($\uparrow$0.70\%). These results confirm that the framework generalizes across graph densities without requiring architectural changes.

\subsection{Transferability of Prompt Optimization}
\label{app:transfer_po}
A natural question is whether routing knowledge learned on one benchmark can transfer to another, potentially avoiding the cost of per-dataset optimization. Table~\ref{tab:transfer} investigates this by applying the router trained on HotpotQA to three unseen target benchmarks: 2Wiki, NewsQA, and NGQA.

The results reveal that cross-dataset routing consistently underperforms in-domain training by a wide margin. Compared to the in-domain EvolveRouter baseline, the transferred PO router suffers F1 drops of $-27.6\%$ on 2Wiki, $-15.0\%$ on NewsQA, and $-28.9\%$ on NGQA. The degradation is particularly severe on NGQA, where the domain-specific reasoning patterns diverge most from HotpotQA. These gaps confirm that learned routing policies are highly task-specific: the router assigns high weight to different agent subsets depending on the benchmark, and these preferences do not generalize across tasks with different reasoning demands. This is precisely why per-dataset training remains necessary---a router optimized on even a closely related multi-hop benchmark fails to capture the agent complementarities specific to the target task.

However, pairing the transferred router with prompts optimized on HotpotQA (PO + PO router) partially recovers the gap on two of the three targets. On NewsQA, F1 increases from 55.05 to 59.87, and on NGQA from 38.43 to 45.13. On 2Wiki the improvement is marginal (56.28 to 57.44). This pattern suggests that while routing weights themselves are non-transferable, prompt refinements carry partial cross-task benefits: better-calibrated agent instructions produce higher-quality outputs regardless of whether the routing policy matches the target task. The recovery is meaningful but incomplete. Even with PO prompts, the transferred router still trails the in-domain baseline by a substantial margin on all three targets, reinforcing that routing and prompt quality contribute complementary gains and that the full benefit of our co-evolution framework requires task-specific optimization of both components.

\begin{table*}[h]
\centering

\resizebox{0.8\textwidth}{!}{
\begin{tabular}{l cc cc cc}
\toprule
\multirow{2}{*}{\textbf{Method}} &
\multicolumn{2}{c}{\textbf{HotpotQA $\rightarrow$2Wiki}} &
\multicolumn{2}{c}{\textbf{HotpotQA $\rightarrow$NewsQA}}&
\multicolumn{2}{c}{\textbf{HotpotQA $\rightarrow$NGQA}} \\
\cmidrule(lr){2-3}\cmidrule(lr){4-5}\cmidrule(lr){6-7}
& \textbf{F1} & \textbf{EM}
& \textbf{F1} & \textbf{EM} 
& \textbf{F1} & \textbf{EM}\\
\toprule
EvolveRouter (in-domain)
&77.70{\tiny$\pm$0.13}  &70.33{\tiny$\pm$0.58} &64.74{\tiny$\pm$0.77}  &41.67{\tiny$\pm$0.47} 
&54.04{\tiny$\pm$0.60}& 9.67{\tiny$\pm$0.58}\\
\midrule
PO + PO router
 &57.44{\tiny$\pm$0.75} &47.00{\tiny$\pm$1.00} &59.87{\tiny$\pm$0.83} &40.33{\tiny$\pm$0.58}
 &45.13{\tiny$\pm$1.47}&4.33{\tiny$\pm$0.58}\\
\rowcolor[RGB]{222,230,241}
\textbf{$\Delta$ vs.\ base (\%)} &$\downarrow$26.08\% &$\downarrow$33.18\% &$\downarrow$7.53\% &$\downarrow$3.22\% & $\downarrow$16.49\%&$\downarrow$55.23\%\\
PO router
&56.28{\tiny$\pm$0.96}&46.33{\tiny$\pm$0.58}
&55.05{\tiny$\pm$0.66}&38.33{\tiny$\pm$0.58}
&38.43{\tiny$\pm$1.64}&4.33{\tiny$\pm$1.15}
\\
\rowcolor[RGB]{222,230,241}
\textbf{$\Delta$ vs.\ base (\%)} &$\downarrow$27.57\% &$\downarrow$34.13\% &$\downarrow$14.97\% &$\downarrow$8.02\% & $\downarrow$28.89\%&$\downarrow$55.23\%
% router
% &58.32&48.00
% &61.24&40.00
% &42.07&6.00
\\
\bottomrule
\end{tabular}
}

% \vspace{0.8em}

% \resizebox{0.8\textwidth}{!}{
% \begin{tabular}{l cc cc cc}
% \toprule
% \multirow{2}{*}{\textbf{Method}} &
% \multicolumn{2}{c}{\textbf{NewsQA$\rightarrow$HotpotQA}} &
% \multicolumn{2}{c}{\textbf{NewsQA$\rightarrow$TriviaQA}}&
% \multicolumn{2}{c}{\textbf{NewsQA$\rightarrow$NGQA}} \\
% \cmidrule(lr){2-3}\cmidrule(lr){4-5}\cmidrule(lr){6-7}
% & \textbf{F1} & \textbf{EM}
% & \textbf{F1} & \textbf{EM} 
% & \textbf{F1} & \textbf{EM}\\
% \toprule
% EvolveRouter (base)
% &72.28{\tiny$\pm$0.86}  &60.33{\tiny$\pm$1.25} &69.83{\tiny$\pm$0.32} &60.00{\tiny$\pm$1.00}
% &54.04{\tiny$\pm$0.60}& 9.67{\tiny$\pm$0.58}\\
% \midrule
% PO + PO router
%  &61.72{\tiny$\pm$1.20} &49.00{\tiny$\pm$1.00} &42.11{\tiny$\pm$0.61} &36.67{\tiny$\pm$0.58}
%  &45.66{\tiny$\pm$0.39}&4.33{\tiny$\pm$0.58}\\
% \rowcolor[RGB]{222,230,241}
% \textbf{$\Delta$ vs.\ base (\%)} &$\downarrow$14.61\% &$\downarrow$18.79\% &$\downarrow$39.70\% &$\downarrow$38.89\% &$\downarrow$15.51\%&$\downarrow$55.23\%\\

% PO router
% &65.77&57.00
% &57.62&44.00
% &45.84&5.00
% \\
% router
% &63.57&50.00
% &53.77&45.00
% &46.48&5.00
% \\
% \bottomrule
% \end{tabular}
% }

\caption{Cross-dataset transferability of routing and prompt optimization. The router is trained on HotpotQA and evaluated on 2Wiki, NewsQA and NGQA.}
\label{tab:transfer}

\end{table*}
\subsection{Ablation study of adaptive K}
Table~\ref{tab:ablation_adaptivek} compares Adaptive K against fixed-K settings across five benchmarks. No single fixed K performs best on all datasets: K=5 is strongest on 2Wiki and HotpotQA, while K=3 edges ahead on NewsQA and NGQA. This confirms that different queries benefit from different collaboration sizes, and a global K is inherently suboptimal. Adaptive K addresses this by determining K per query through answer agreement, matching or exceeding the best fixed-K on every benchmark. Compared to the KG Router base, Adaptive K improves F1 by +1.52\% on 2Wiki, +1.53\% on HotpotQA, +1.25\% on NewsQA, and +0.70\% on NGQA, with a notably larger EM gain of +6.51\% on TriviaQA. 

Table~\ref{tab:ablation_adaptivek_saving} reports the average number of agents queried by Adaptive K. Across benchmarks, the average K ranges from 2.9 to 7.4, yielding a 69–88\% reduction in agent calls compared to the full ensemble (K=24). The savings are largest on multi-hop benchmarks (2Wiki and HotpotQA), where top-ranked agents tend to agree quickly, and smallest on TriviaQA, where answer diversity requires consulting more agents before reaching consensus.

\begin{table*}[t]
\centering
\resizebox{\textwidth}{!}{
\begin{tabular}{l cc cc cc cc cc}
\toprule
\multirow{2}{*}{\textbf{Method}} &
\multicolumn{2}{c}{\textbf{2Wiki}} &
\multicolumn{2}{c}{\textbf{HotpotQA}} &
\multicolumn{2}{c}{\textbf{NewsQA}} &
\multicolumn{2}{c}{\textbf{TriviaQA}} &
\multicolumn{2}{c}{\textbf{NGQA}} \\
\cmidrule(lr){2-3}\cmidrule(lr){4-5}\cmidrule(lr){6-7}\cmidrule(lr){8-9}\cmidrule(lr){10-11}
& \textbf{F1} & \textbf{EM}
& \textbf{F1} & \textbf{EM}
& \textbf{F1} & \textbf{EM}
& \textbf{F1} & \textbf{EM}
& \textbf{F1} & \textbf{EM} \\
\toprule
KG Router (base)
&75.54{\tiny$\pm$0.44}&67.67{\tiny$\pm$0.58}
&69.35{\tiny$\pm$0.56}&56.33{\tiny$\pm$0.58}
&63.02{\tiny$\pm$2.44}&41.33{\tiny$\pm$0.58}
&62.75{\tiny$\pm$3.69}&51.33{\tiny$\pm$2.51}
&49.95{\tiny$\pm$1.29}&8.67{\tiny$\pm$1.53}
\\
\rowcolor{gray!20}
\multicolumn{11}{c}{\textit{Adaptive K Ablation}} \\
Fixed K=3
&75.49{\tiny$\pm$0.59}&66.67{\tiny$\pm$0.58}
&67.08{\tiny$\pm$1.17}&54.00{\tiny$\pm$1.00}
&60.92{\tiny$\pm$0.68}&40.67{\tiny$\pm$0.58}
&61.60{\tiny$\pm$0.86}&51.00{\tiny$\pm$1.00}
&49.76{\tiny$\pm$0.49}&8.67{\tiny$\pm$0.58}
\\
Fixed K=5
&76.22{\tiny$\pm$1.46}&67.00{\tiny$\pm$1.00}
&69.83{\tiny$\pm$1.36}&56.67{\tiny$\pm$0.58}
&62.83{\tiny$\pm$0.46}&41.00{\tiny$\pm$1.00}
&62.60{\tiny$\pm$0.81}&52.33{\tiny$\pm$0.58}
&48.94{\tiny$\pm$0.59}&7.33{\tiny$\pm$0.58}
\\
Fixed K=10
&76.16{\tiny$\pm$0.81}&67.00{\tiny$\pm$1.00}
&69.40{\tiny$\pm$0.88}&56.00{\tiny$\pm$1.00}
&61.08{\tiny$\pm$1.38}&41.00{\tiny$\pm$1.00}
&61.18{\tiny$\pm$0.76}&49.67{\tiny$\pm$1.15}
&48.34{\tiny$\pm$0.31}&7.00{\tiny$\pm$0.00}
\\
\midrule
+ Adaptive K
&76.69{\tiny$\pm$0.31}&68.00{\tiny$\pm$0.00}
&70.41{\tiny$\pm$0.50}&57.00{\tiny$\pm$0.58}
&63.81{\tiny$\pm$0.83}&41.33{\tiny$\pm$0.58}
&63.10{\tiny$\pm$0.51}&54.67{\tiny$\pm$0.58}
&50.30{\tiny$\pm$1.74}&9.00{\tiny$\pm$1.00}
\\
\rowcolor[RGB]{222,230,241}
\textbf{$\Delta$ vs.\ base (\%)} 
&$\uparrow$ 1.52\%&$\uparrow$ 0.49\%&$\uparrow$ 1.53\%&$\uparrow$ 1.19\%&$\uparrow$ 1.25\%&$\uparrow$ 0.00\%&$\uparrow$ 0.56\%&$\uparrow$ 6.51\%&$\uparrow$ 0.70\%&$\uparrow$ 3.81\%
\\

\bottomrule
\end{tabular}
}
\caption{Ablation study of adaptive K across five benchmarks. Adaptive K exceeds the best fixed-K setting on every dataset.}
\label{tab:ablation_adaptivek}
\end{table*}

\begin{table*}[t]
\centering
\resizebox{0.65\textwidth}{!}{
\begin{tabular}{l ccccc}
\toprule
\textbf{Method} & \textbf{2Wiki} & \textbf{HotpotQA} & \textbf{NewsQA} & \textbf{TriviaQA} & \textbf{NGQA} \\
\toprule
Full ensemble
&24&24&24&24&24
\\
Adaptive Avg. K
&2.9&2.9&4.2&7.4&3.3
\\
\rowcolor[RGB]{222,230,241}
\textbf{Savings (\%)} &88\% &88\% &83\% &69\% &86\% \\
\bottomrule
\end{tabular}
}
\caption{Average number of agents queried by Adaptive K compared to the full ensemble (K=24). Savings indicate the percentage reduction in agent calls.}
\label{tab:ablation_adaptivek_saving}
\end{table*}
\section{Prompt Design}
\label{app:prompts}
We provide the complete set of prompts used in our framework for reproducibility. Our prompt architecture consists of three layers. First, each of the six agent roles (Raw, CoT, Self-Consistency, Multi-Agent Debate, ReAct-Reflection, and Multi-Agent Summary) is defined by a role-specific prompt that encodes its reasoning strategy (Figure~\ref{fig:prompt1} --~\ref{fig:prompt2}). Second, a shared system wrapper standardizes the output format and enforces answer constraints across all agents (Figure~\ref{fig:prompt-system}). At inference time, the role prompt is injected into the system wrapper's \textless role\_prompt\textgreater\space slot, ensuring that all agents share the same output conventions while maintaining distinct reasoning behaviors. Third, when prompt refinement is triggered during co-evolution, the rewriter LLM receives a structured template (Figure~\ref{fig:prompt-rewrite}) containing the current prompt, aggregated error patterns from router diagnostics, and representative failure cases. The rewriter is constrained to make minimal, targeted edits while preserving the role's core strategy and output format. 

\section{Case Study}
\label{app:case}
We present two types of case studies to illustrate the behavior of EvolveRouter.

\paragraph{Agent routing examples.} Figure~\ref{fig:routing-examples} shows per-question routing snapshots for three representative queries. For each query, we display the top-4 and bottom-4 agents ranked by router weight, along with their generated answers. These examples illustrate that the router consistently assigns high weights to agents that produce correct answers while down-weighting agents that generate incorrect or noisy outputs.

\paragraph{Prompt optimization case studies.}
Figures~\ref{fig:po-case-1} --~\ref{fig:po-case-3} present three questions where prompt refinement flipped the prediction from incorrect to correct. Each example highlights a distinct failure mode. In Figure~\ref{fig:po-case-1}, the agent returned a related entity instead of the queried attribute (entity–attribute confusion). In Figure~\ref{fig:po-case-2}, agents misinterpreted an entity question as a yes/no question and returned "yes" instead of extracting the answer. In Figure~\ref{fig:po-case-3}, agents selected a contextually plausible but incorrect entity from multiple candidates in the passage. In all three cases, the refined prompts introduced targeted reasoning instructions, such as explicit entity tracing and answer precision checks, that corrected the failure without altering the agent's core strategy.
\begin{figure*}[t]
\centering
\begin{promptbox}{Agent Role Prompts --- Part I}
\small
\textbf{raw:}\\
Given a question, a news context, and retrieved documents, answer the question.

The final answer must appear on the last line in the format: \boxed{<answer>}.

\vspace{6pt}
\textbf{cot (Chain-of-Thought):}\\
You are a multi-hop reasoning expert and an expert QA agent. Given a question, and the context, think step-by-step. 

The final answer must appear on the last line in the format: \boxed{<answer>}.

\vspace{6pt}

\textbf{mad (Multi-Agent Debate):}\\
\textbf{debate\_debater\_a:}\\
You are Debater A. Your goal is to propose the most plausible answer using the provided context.
\begin{itemize}
    \item Make ONE clear claim (the candidate answer).
    \item Support it with 1–2 ultra-short quotes (verbatim substrings) and name the hops.
    \item Explain the link between the quotes in $\leq$ 2 sentences.
\end{itemize}
Do NOT use outside knowledge and do NOT output the final boxed answer. Make your answer really short and concise.

\textbf{debate\_debater\_b:}\\
You are Debater B. Your goal is to stress-test A’s claim using ONLY the provided context.
\begin{itemize}
    \item If A’s quotes or hops are weak, inconsistent, or incomplete, point it out and give corrected quotes/hops.
    \item If a better candidate exists, state ONE alternative with 1–2 short quotes and $\leq$ 2 sentences of reasoning.
    \item If A is already well-supported, briefly confirm but add one missing check.
\end{itemize}
Do NOT use outside knowledge and do NOT output the final boxed answer. Make your answer really short and concise. 

\textbf{debate\_judge:}\\
You are the Judge. Read A and B as supporting analyses and decide the best final answer using ONLY the given context. If evidence is thin, still make your best context-based guess.

Output MUST include nothing but brief final answer in the format: \boxed{<answer>}.

\vspace{6pt}
\textbf{react\_reflect (ReAct-Reflection):}\\
\textbf{react:}\\
You are a multi-hop reasoning expert and an expert QA agent. Given a question, a news context, and retrieved documents, think step-by-step, silently chain facts to derive a thinking plan, then use this plan to derive the final brief answer.

Your output format MUST be a brief final answer on the last line in the format: \boxed{<answer>}.

\textbf{reflect:}\\
You are a judge overseeing a multi-hop reasoning expert and an expert QA agent. Given a question, a news context, and retrieved documents, you will evaluate the agent’s answer based on the correctness and notes. If the answer is incorrect or incomplete, provide constructive feedback and suggest specific revisions to improve the answer. If the answer is correct and complete, indicate that no further revisions are needed.

Your output MUST end with either:
\begin{itemize}
    \item “Status: revise” followed by specific feedback and revision suggestions, if the answer needs improvement.
    \item “Status: final” if the answer is correct and complete.
\end{itemize}
If you indicate “Status: revise”, also include a short “Feedback: \textless your feedback here\textgreater” section before the final answer.
\end{promptbox}
\caption{Agent role prompts (Part I) for multi-hop QA}
\label{fig:prompt1}
\end{figure*}

\begin{figure*}[t]
\centering
\begin{promptbox}{Agent Role Prompts --- Part II}
\small
\textbf{sc (Self-Consistency):}\\
You are a self-consistency agent. Independently sample multiple plausible entity selections for the given question and context, then internally perform majority voting to decide the final set. Generate diverse candidate sets internally, then pick the majority-agreed entities.

The final answer must appear on the last line in the format: \boxed{<answer>}.

\vspace{6pt}

\textbf{think:}\\
You are a multi-hop reasoning expert and an expert QA agent. \\Given a question, a news context, and retrieved documents, think step-by-step, chain facts to derive the answer.\\Give your final answer as a single entity, and a concise reasoning process that leads to the answer.

\vspace{6pt}
\textbf{summarize:}\\
You are the multi-hop reasoning expert and an expert QA agent. You receive outputs from other agents. Use them as \textbf{supporting signals}.\\If A and B agree on the same short span, return it. If they differ, pick the best answer with your own reasoning.\\Your output format MUST end with the brief final answer on the last line in the format: \boxed{<answer>}.
\end{promptbox}
\caption{Agent role prompts (Part II) for multi-hop QA}
\label{fig:prompt2}
\end{figure*}

\begin{figure*}[t]
\centering
\begin{promptbox}{System Wrapper and Output Constraints}
\small
\textbf{System prompt (applied to all agents):}\\
You are a QA agent. Your role is: \texttt{<role\_name>}. Your way of thinking and acting is as follows: \texttt{<role\_prompt>}. Given a question and context, output ONLY one JSON object: \{"answer": "\textless short factual answer\textgreater"\}.
\begin{itemize}
\item The answer must be a short factual string (yes/no or a name/date/number from context).
\item NEVER explain. NEVER return empty. No sentences or verbs.
\item Keep the answer $\leq$ 12 words.
\end{itemize}
\vspace{6pt}
\textbf{Output format rules:}\\
Here are the rules you must \textbf{STRICTLY} follow:
\begin{enumerate}
\item Always return the answer as the SHORTEST exact entity only. The answer is always within 10 words, and usually within 5 words.
\item If the question is yes/no, respond strictly with \emph{yes} or \emph{no} only.
\item For year ranges, never use hyphens; instead, use "from XXXX to YYYY" or "XXXX until YYYY".
\item Do not output sentences, explanations, or phrases with verbs; the answer must be a single entity expression only.
\item One way or another, you must return your best guess, and the final answer must be in the format: \boxed{<answer>}.

\end{enumerate}
\end{promptbox}
\caption{System-level wrapper and output constraints applied to every agent call. The role prompt from Figures~\ref{fig:prompt1}--\ref{fig:prompt2} is injected into the \texttt{<role\_prompt>} slot.}
\label{fig:prompt-system}
\end{figure*}

\begin{figure*}[t]
\centering
\begin{promptbox}{Rewrite System Prompt}
\small
You are an expert prompt engineer for multi-hop question answering agents.\\
Your task: Given a current prompt for a QA agent role, failure examples, and error patterns, produce an \textbf{IMPROVED} version of the prompt that addresses the identified weaknesses.
\vspace{6pt}

\textbf{Mutation strategy} --- Make only minimal, targeted changes:
\begin{itemize}
\item You may ADD 1--3 sentences of clarifying instructions.
\item You may REPHRASE existing sentences for clarity.
\item You may REMOVE unhelpful or misleading instructions.
\item You must NOT rewrite the entire prompt from scratch or change the overall structure.
\item Keep the same opening sentence and general flow.
\end{itemize}

\textbf{Guidelines:}
\begin{itemize}
\item Keep the role's core strategy (e.g., CoT stays CoT, self-consistency stays self-consistency).
\item Preserve the output format: the agent must return a JSON object \{"answer": \textless short factual answer\textgreater''\}.
\item Add specific, actionable reasoning instructions based on the failure patterns. For example:
  \begin{itemize}
  \item If the agent gives partial answers: ``Extract ALL entities mentioned in the question and trace each one through the context before answering.''
  \item If the agent answers with wrong entities: ``Double-check that your answer directly answers the specific question being asked, not a related but different question.''
  \item If the agent gives empty/unanswerable responses: ``You MUST always provide your best guess. Never say unanswerable or N/A.''
  \end{itemize}
\item Keep the prompt concise (under 500 words). Longer is not better.
\item Focus on the 1--2 most impactful changes based on the failure patterns.
\vspace{6pt}

Output: Return ONLY the improved prompt text. No explanation, no markdown, no quotes.

\end{itemize}

\end{promptbox}
\caption{Input template for the rewriter LLM. The template provides the current prompt, aggregated error patterns from router diagnostics, and representative failure cases with their F1 scores to guide targeted prompt refinement.}
\label{fig:prompt-rewrite}
\end{figure*}

\begin{figure*}[t]
\centering
\begin{promptbox2}{Agent Routing Examples}
\small
\textbf{Example 1} \\
\textbf{Question:} Were Scott Derrickson and Ed Wood of the same nationality? \\
\textbf{Gold Answer:} yes \\
\textbf{Model Prediction:} yes

\vspace{4pt}
\textbf{Top-4 Agents} \\
{\footnotesize
\texttt{BACKBONE::gpt\_oss\_20b::AGENT::raw} (0.092484698) $\to$ \textit{yes} \\
\texttt{BACKBONE::gpt\_oss\_20b::AGENT::cot} (0.089811556) $\to$ \textit{yes} \\
\texttt{BACKBONE::gpt\_oss\_20b::AGENT::sc} (0.069166116) $\to$ \textit{yes} \\
\texttt{BACKBONE::gpt\_oss\_20b::AGENT::mad} (0.064756475) $\to$ \textit{yes}
}

\textbf{Last-4 Agents} \\
{\footnotesize
\texttt{BACKBONE::mixtral\_8x7b::AGENT::summary} (0.006042599) $\to$ \textit{American} \\
\texttt{BACKBONE::qwen2p5\_7b\_turbo::AGENT::summary} (0.007424950) $\to$ \textit{no} \\
\texttt{BACKBONE::llama3\_8b\_lite::AGENT::summary} (0.009983738) $\to$ \textit{American} \\
\texttt{BACKBONE::mixtral\_8x7b::AGENT::mad} (0.027359875) $\to$ \textit{American}
}

\vspace{6pt}\hrule\vspace{6pt}
\textbf{Example 2} \\
\textbf{Question:} What is the middle name of the actress who plays Bobbi Bacha in Suburban Madness? \\
\textbf{Gold Answer:} Ann \\
\textbf{Model Prediction:} Ann

\vspace{4pt}
\textbf{Top-4 Agents} \\
{\footnotesize
\texttt{BACKBONE::gpt\_oss\_20b::AGENT::raw} (0.097907) $\to$ \textit{Ann} \\
\texttt{BACKBONE::gpt\_oss\_20b::AGENT::cot} (0.091182) $\to$ \textit{Ann} \\
\texttt{BACKBONE::gpt\_oss\_20b::AGENT::sc} (0.068578) $\to$ \textit{Ann} \\
\texttt{BACKBONE::llama3\_8b\_lite::AGENT::raw} (0.066690) $\to$ \textit{Bacha}
}

\textbf{Last-4 Agents} \\
{\footnotesize
\texttt{BACKBONE::mixtral\_8x7b::AGENT::summary} (0.004486) $\to$ \textit{Sela} \\
\texttt{BACKBONE::qwen2p5\_7b\_turbo::AGENT::summary} (0.010262) $\to$ \textit{Sela} \\
\texttt{BACKBONE::llama3\_8b\_lite::AGENT::summary} (0.013433) $\to$ \textit{Ward} \\
\texttt{BACKBONE::qwen2p5\_7b\_turbo::AGENT::mad} (0.029658) $\to$ \textit{Sela}
}

\vspace{6pt}\hrule\vspace{6pt}
\textbf{Example 3} \\
\textbf{Question:} What happened to the bus? \\
\textbf{Gold Answer:} rolled over \\
\textbf{Model Prediction:} rolled over

\vspace{4pt}
\textbf{Top-4 Agents} \\
{\footnotesize
\texttt{BACKBONE::llama3\_8b\_lite::AGENT::sc} (0.068562) $\to$ \textit{rolled over} \\
\texttt{BACKBONE::llama3\_8b\_lite::AGENT::mad} (0.066393) $\to$ \textit{rolled over} \\
\texttt{BACKBONE::llama3\_8b\_lite::AGENT::cot} (0.064613) $\to$ \textit{bus rolled over} \\
\texttt{BACKBONE::gpt\_oss\_20b::AGENT::cot} (0.064280) $\to$ \textit{crash}
}

\textbf{Last-4 Agents} \\
{\footnotesize
\texttt{BACKBONE::mixtral\_8x7b::AGENT::sc} (0.025558) $\to$ \textit{Bus crashed} \\
\texttt{BACKBONE::mixtral\_8x7b::AGENT::raw} (0.028061) $\to$ \textit{Bus crashed in Utah} \\
\texttt{BACKBONE::qwen2p5\_7b\_turbo::AGENT::sc} (0.031285) $\to$ \textit{left a highway and rolled over} \\
\texttt{BACKBONE::mixtral\_8x7b::AGENT::mad} (0.031547) $\to$ \textit{Bus crashed in Utah}
}

\end{promptbox2}
\caption{Per-question agent routing snapshots for three examples. Each case shows the question, gold answer, model prediction, and the four highest- and lowest-probability agents with their generated answers.}
\label{fig:routing-examples}
\end{figure*}

\begin{figure}[t]
\centering
\begin{promptbox2}{Case Study 1: Entity–Attribute Confusion}
\small
\textbf{Question:} What was the father of Kasper Schmeichel voted to be by the IFFHS in 1992? \\
\textbf{Gold Answer:} World's Best Goalkeeper
\vspace{3pt}

\textbf{Before PO} \hfill \textcolor{red!70!black}{F1 = 0.000}\\
{\footnotesize Final prediction: \textit{Peter Schmeichel} \quad (confused the entity asked about with the person)}
\vspace{3pt}

\textbf{After PO} \hfill \textcolor{green!50!black}{F1 = 1.000}\\
{\footnotesize Final prediction: \textit{World's Best Goalkeeper}}
\vspace{3pt}

{\footnotesize
\textbf{What changed:} The refined CoT prompt added explicit instructions to "extract ALL entities mentioned in the question and trace each one through the context before answering" and to "double-check that your answer directly addresses the specific question being asked." With the base prompt, agents fixated on the person entity (Peter Schmeichel) rather than the distinction he was voted for. After PO, the top-ranked agents correctly distinguished the entity being asked about from the entity providing context.
}
\end{promptbox2}
\caption{PO case study 1: entity–attribute confusion. }
\label{fig:po-case-1}
\end{figure}

\begin{figure}[t]
\centering
\begin{promptbox2}{Case Study 2: Question Type Misidentification}
\small
\textbf{Question:} Kaiser Ventures corporation was founded by an American industrialist who became known as the father of modern American shipbuilding? \\
\textbf{Gold Answer:} Henry J. Kaiser
\vspace{3pt}

\textbf{Before PO} \hfill \textcolor{red!70!black}{F1 = 0.000}\\
{\footnotesize Final prediction: \textit{yes} \quad (misinterpreted as a yes/no question)}
\vspace{3pt}

\textbf{After PO} \hfill \textcolor{green!50!black}{F1 = 1.000}\\
{\footnotesize Final prediction: \textit{Henry J. Kaiser}}
\vspace{3pt}

{\footnotesize
\textbf{What changed:} Before PO, most agents treated this as a yes/no verification question and returned "yes" or "no." The refined SC prompt introduced an explicit "Entity Extraction" step that requires identifying all entities before answering and an "Answer Precision" step that instructs to "provide the full and exact answer, not a partial or generalized term." After PO, the SC and MAD agents correctly extracted the named entity, shifting the weighted vote from "yes" to "Henry J. Kaiser."
}
\end{promptbox2}
\caption{PO case study 2: question type misidentification.}
\label{fig:po-case-2}
\end{figure}

\begin{figure}[t]
\centering
\begin{promptbox2}{Case Study 3: Distractor Entity Selection}
\small
\textbf{Question:} Alvaro Mexia had a diplomatic mission with which tribe of indigenous people? \\
\textbf{Gold Answer:} Apalachees
\vspace{3pt}

\textbf{Before PO} \hfill \textcolor{red!70!black}{F1 = 0.000}\\
{\footnotesize Final prediction: \textit{Seminole} \quad (extracted a related but incorrect tribe)}
\vspace{3pt}

\textbf{After PO} \hfill \textcolor{green!50!black}{F1 = 1.000}\\
{\footnotesize Final prediction: \textit{Apalachees}}
\vspace{3pt}

{\footnotesize
\textbf{What changed:} The context mentions multiple indigenous tribes, and before PO the agents predominantly returned ``Seminole''---a tribe present in the passage but not the one involved in the diplomatic mission. After PO, the refined prompts' emphasis on tracing entity relationships step-by-step led agents to follow the reasoning chain from ``Alvaro Mexia'' through ``diplomatic mission'' to the correct tribe.
}
\end{promptbox2}
\caption{PO case study 3: distractor entity selection. }
\label{fig:po-case-3}
\end{figure}

\end{document}